\documentclass{article}

\usepackage{fullpage}
\usepackage{url,hyperref,lineno,microtype,subcaption}
\usepackage{booktabs} 
\usepackage{graphicx}				
\usepackage{float}
\usepackage{amssymb,amsmath,mathtools}

\newtheorem{execexample}{\bf Execution Example}

\title{Non-monotonic Logical Reasoning Guiding Deep Learning for
  Explainable Visual Question Answering}

\author{  Heather Riley\\
  Electrical and Computer Engineering\\
  University of Auckland, NZ \\
  \texttt{hril230@aucklanduni.ac.nz}
  \and
  Mohan Sridharan\\
  School of Computer Science\\
  University of Birmingham, UK\\
  \texttt{m.sridharan@bham.ac.uk}
}

\date{}

\begin{document}

\maketitle


\begin{abstract}
  State of the art algorithms for many pattern recognition problems
  rely on deep network models. Training these models requires a large
  labeled dataset and considerable computational resources. Also, it
  is difficult to understand the working of these learned models,
  limiting their use in some critical applications. Towards addressing
  these limitations, our architecture draws inspiration from research
  in cognitive systems, and integrates the principles of commonsense
  logical reasoning, inductive learning, and deep learning. In the
  context of answering explanatory questions about scenes and the
  underlying classification problems, the architecture uses deep
  networks for extracting features from images and for generating
  answers to queries.  Between these deep networks, it embeds
  components for non-monotonic logical reasoning with incomplete
  commonsense domain knowledge, and for decision tree induction. It
  also incrementally learns and reasons with previously unknown
  constraints governing the domain's states. We evaluated the
  architecture in the context of datasets of simulated and real-world
  images, and a simulated robot computing, executing, and providing
  explanatory descriptions of plans.  Experimental results indicate
  that in comparison with an ``end to end'' architecture of deep
  networks, our architecture provides better accuracy on
  classification problems when the training dataset is small,
  comparable accuracy with larger datasets, and more accurate answers
  to explanatory questions. Furthermore, incremental acquisition of
  previously unknown constraints improves the ability to answer
  explanatory questions, and extending non-monotonic logical reasoning
  to support planning and diagnostics improves the reliability and
  efficiency of computing and executing plans on a simulated robot.
\end{abstract}

\section{Introduction}
\label{sec:introduction}
Deep neural network architectures and the associated algorithms
represent the state of the art for many perception and control
problems in which their performance often rivals that of human
experts. These architectures and algorithms are increasingly being
used for a variety of tasks such as object recognition, gesture
recognition, object manipulation, and obstacle avoidance, in domains
such as healthcare, surveillance, and navigation. Common limitations
of deep networks are that they are computationally expensive to train,
and require a large number of labeled training samples to learn an
accurate mapping between input(s) and output(s) in complex domains. It
is not always possible to satisfy these requirements, especially in
dynamic domains where previously unseen situations often change the
mapping between inputs and outputs over time. Also, it is challenging
to understand or provide an explanatory description of the observed
behavior of a learned deep network model.  Furthermore, it is
difficult to use domain knowledge to improve the computational
efficiency of learning these models or the reliability of the
decisions made by these models. Consider a self-driving car on a busy
road. Any error made by the car, e.g., in recognizing or responding to
traffic signs, can result in serious accidents and make humans more
reluctant to use such cars.  In general, it is likely that humans
interacting with a system designed for complex domains, with autonomy
in some components, will want to know why and how the system arrived
at particular conclusions; this ``explainability'' will help designers
improve the underlying algorithms and their performance.
Understanding the operation of these systems will also help human
users build trust in the decisions made by these systems. Despite
considerable research in recent years, providing explanatory
descriptions of decision making and learning continues to be an open
problem in AI.

We consider Visual Question Answering (VQA) as a motivating example of
a complex task that inherently requires explanatory descriptions of
reasoning and learning. Given a scene and a natural language question
about images of the scene, the objective of VQA is to provide an
accurate answer to the question. These questions can be about the
presence or absence of particular objects in the image, the
relationships between objects, or the potential outcome of executing
particular actions in the scene. For instance, a system recognizing
and responding to traffic signs on a self-driving car may be posed
questions such as ``what is the traffic sign in the image?'', or
``what is the meaning of this traffic sign?'', and a system
controlling a robot arm constructing stable arrangements of objects on
a tabletop may be asked ``why is this structure unstable?''  or ``what
would make the structure stable?''.  We assume that any such questions
are provided as (or transcribed into) text, and that answers to
questions are also generated as text (that may be converted to speech)
using existing software. Deep networks represent the state of the art
for VQA, but are characterized by the known limitations described
above. We seek to address these limitations by drawing inspiration
from research in cognitive systems, which indicates that reliable,
efficient, and explainable reasoning and learning can be achieved in
complex problems by jointly reasoning with commonsense domain
knowledge and learning from experience.  Specifically, the
architecture described in this paper tightly couples knowledge
representation, reasoning, and learning, and exploits the
complementary strengths of deep learning, inductive learning, and
non-monotonic logical reasoning with incomplete commonsense domain
knowledge.  We describe the following characteristics of the
architecture:
\begin{itemize}
\item For any input image of a scene of interest, Convolutional Neural
  Networks (CNNs) extract concise visual features characterizing the
  image.

\item Non-monotonic logical reasoning with the extracted features and
  incomplete commonsense domain knowledge is used to classify the
  input image, and to provide answers to explanatory questions about
  the classification and the scene.

\item Feature vectors that the non-monotonic logical reasoning is
  unable to classify are used to train a decision tree classifier that
  is also used to answer questions about the classification during
  testing.

\item Feature vectors not classified by non-monotonic logical
  reasoning, along with the output of the decision tree classifier,
  train a Recurrent Neural Network (RNN) that is used to answer
  explanatory questions about the scene during testing.

\item Feature vectors not classified by non-monotonic logical
  reasoning are also used to inductively learn, and subsequently
  reason with, constraints governing domain states; and 

\item Reasoning with commonsense knowledge is expanded (when needed)
  to support planning, diagnostics, and the ability to answer related
  explanatory questions.
\end{itemize}
This architecture builds on our prior work on combining commonsense
inference with deep learning~\cite{riley:hai18,mota:rss19} by
introducing the ability to learn and reason with constraints governing
domain states, and extending reasoning with commonsense knowledge to
support planning and diagnostics to achieve any given goal.

Although we use VQA as a motivating example, it is not the main focus
of our work. State of the art algorithms for VQA focus on generalizing
to images from different domains, and are evaluated on benchmark
datasets of several thousand images drawn from different
domains~\cite{shrestha:cvpr19}. Our focus, on the other hand, is on
transparent reasoning and learning in any given domain in which a
large, labeled dataset is not readily available. Towards this
objective, our approach explores the interplay between non-monotonic
logical reasoning, incremental inductive learning, and deep learning.
We thus neither compare our architecture and algorithms with state of
the art algorithms for VQA, nor use large benchmark VQA datasets for
evaluation.  Instead, we evaluate our architecture's capabilities in
the context of: (i) estimating the stability of configurations of
simulated blocks on a tabletop; (ii) recognizing different traffic
signs in a benchmark dataset of images; and (iii) a simulated robot
delivering messages to the intended recipients at different locations.
The characteristics of these tasks and domains match our objective. In
both domains, we focus on answering explanatory questions about images
of scenes and the underlying classification problems (e.g.,
recognizing traffic signs).  In addition, we demonstrate how our
architecture can be adapted to enable a robot assisting humans to
compute and execute plans, and to answer questions about these plans.
Experimental results show that in comparison with an architecture
based only on deep networks, our architecture provides: (i) better
accuracy on classification problems when the training dataset is
small, and comparable accuracy on larger datasets; and (ii)
significantly more accurate answers to explanatory questions about the
scene.  We also show that the incremental acquisition of state
constraints improves the ability to answer explanatory questions, and
to compute minimal and correct plans.

We begin with a discussion of related work in
Section~\ref{sec:relwork}. The architecture and its components are
described in Section~\ref{sec:arch}, with the experimental results
discussed in Section~\ref{sec:expresults}.
Section~\ref{sec:conclusions} then describes the conclusions and
directions for further research.


\section{Related work}
\label{sec:relwork}
State of the art approaches for VQA are based on deep learning
algorithms~\cite{jiang:TR15,malinowski:IJCV17,masuda:cvpr16,pandhre:TR17,shrestha:cvpr19,zhang:TR17}.
These algorithms use labeled data to train neural network
architectures with different arrangements of layers and connections
between them, capturing the mapping between the inputs (e.g., images,
text descriptions) and the desired outputs (e.g., class labels, text
descriptions). Although deep networks have demonstrated the ability to
model complex non-linear mappings between inputs and outputs for
different pattern recognition tasks, they are computationally
expensive and require large, labeled training datasets. They also make
it difficult to understand and explain the internal representations,
identify changes that will improve performance, or to transfer
knowledge acquired in one domain to other related domains. In
addition, it is challenging to accurately measure performance or
identify dataset bias, e.g., deep networks can answer questions about
images using question-answer training samples without even reasoning
about the images~\cite{jabri:eccv16,teney:TR16,zhang:TR17}.  There is
on-going research on each of these issues, e.g., to explain the
operation of deep networks, reduce training data requirements and
bias, reason with domain knowledge, and incrementally learn the domain
knowledge. We review some of these approaches below, primarily in the
context of VQA.

Researchers have developed methods to understand the internal
reasoning of deep networks and other machine learning algorithms.
Selvaraju et al.~\cite{selvaraju:iccv17} use the gradient in the last
convolutional layer of a CNN to compute the relative contribution
(importance weight) of each neuron to the classification decision
made. However, the weights of neurons do not provide an intuitive
explanation of the CNN's operation or its internal representation.
Researchers have also developed general approaches for understanding
the predictions of any given machine learning algorithm. For instance,
Koh and Liang~\cite{koh:icml17} use second-order approximations of
influence diagrams to trace any model's prediction through a learning
algorithm back to the training data in 4order to identify training
samples most responsible for any given prediction.  Ribeiro et
al.~\cite{ribeiro:kdd16} developed a framework that analyzes any
learned classifier model by constructing a interpretable simpler model
that captures the essence of the learned model. This framework
formulates the task of explaining the learned model, based on
representative instances and explanations, as a submodular
optimization problem. In the context of VQA, Norcliffe et
al.~\cite{norcliffe:nips18} provide interpretability by introducing
prior knowledge of scene structure as a graph that is learned from
observations based on the question under consideration.  Object
bounding boxes are graph nodes while edges are learned using an
attention model conditioned on the question. Mascharka et
al.~\cite{mascharka:cvpr18} augment a deep network architecture with
an image-space attention mechanism based on a set of composable visual
reasoning primitives that help examine the intermediate outputs of
each module. Li et al.~\cite{li:corr18} introduce a captioning model
to generate an image's description, reason with the caption and the
question to construct an answer, and use the caption to explain the
answer. However, these algorithms do not support the use of
commonsense reasoning to (i) provide meaningful explanatory
descriptions of learning and reasoning; (ii) guide learning to be more
efficient; or (iii) provide reliable decisions when large training
datasets are not available.


The training data requirements of a deep network can be reduced by
directing attention to data relevant to the tasks at hand. In the
context of VQA, Yang et al.~\cite{yang:cvpr16} use a Long Short-Term
Memory (LSTM) network to map the question to an encoded vector,
extract a feature map from the input image using a CNN, and use a
neural network to compute weights for feature vectors based on their
relevance to the question. A stacked attention network is trained to
map the weighted feature vectors and question vector to the answer,
prioritizing feature vectors with greater weights.  Schwartz et
al.~\cite{schwartz:nips17} use learned higher-order correlations
between various data modalities to direct attention to elements in the
data modalities that are relevant to the task at hand.  Lu et
al.~\cite{lu:nips16} use information from the question to identify
relevant image regions and uses information from the image to identify
relevant words in the question. A co-attentional model jointly and
hierarchically reasons about the image and the question at three
levels, embedding words in a vector space, using one-dimensional CNNs
to model information at the phrase level, and using RNNs to encode the
entire question. A generalization of this work, a Bilinear Attention
Network, considers interactions between all region proposals in the
image with all words in the (textual) question~\cite{kim:nips18}. A
Deep Attention Neural Tensor Network for VQA, on the other hand, uses
tensor-based representations to discover joint correlations between
images, questions, and answers~\cite{bai:eccv18}. The attention module
is based on a discriminative reasoning process, and regression with
KL-divergence losses improves scalability of training and convergence.
Recent work by combines top-down and bottom-up attention mechanisms,
with the top-down mechanism providing an attention distribution over
object proposals provided by the bottom-up
mechanism~\cite{anderson:cvpr18}.

In addition to reducing the training data requirements, researchers
have focused on reducing the number of annotated samples needed for
training, and on minimizing the bias in deep network models. In the
context of VQA, Lin et al.~\cite{lin:eccv14} iteratively revise a
model trained on an initial training set by expanding the training set
with image-question pairs involving concepts it is uncertain about,
with an ``oracle'' (human annotater) providing the answers.  This
approach reduces annotation time, but the database includes just as
many images and questions as before. Goyal et al.~\cite{goyal:cvpr17}
provide a balanced dataset with each question associated with a pair
of images that require different answers, and provide a counterexample
based explanation for each image-question pair.  Agrawal et
al.~\cite{agrawal:cvpr18}, on the other hand, separate the recognition
of visual concepts in an image from the identification of an answer to
any given question, and include inductive biases to prevent the
learned model from relying predominantly on priors in the training
data.

In computer vision, robotics and other applications, learning from
data can often be made more efficient by reasoning with prior
knowledge about the domain. In the context of VQA, Wang et
et.~\cite{wang:ijcai17} reason with knowledge about scene objects to
answer common questions about these objects, significantly expanding
the range of natural language questions that can be answered without
making the training data requirements impractical. However, this
approach does not reduce the amount of data required to train the deep
network. Furbach et al.~\cite{furbach:KI10} directly use a knowledge
base to answer questions and do not consider the corresponding images
as inputs. Wagner et al.~\cite{wagner:eccvwrkshp18}, on the other
hand, use physics engines and prior knowledge of domain objects to
realistically simulate and explore different situations. These
simulations guide the training of deep network models that anticipate
action outcomes and answer questions about all situations. Based on
the observation that VQA often requires reasoning over multiple steps,
Wu et al.~\cite{wu:nips18} construct a chain of reasoning for
multi-step and dynamic reasoning with relations and objects. This
approach iteratively forms new relations between objects using
relational reasoning operations, and forms new compound objects using
object refining operations, to improve VQA performance. Given the
different components of a VQA system, Teney and van den
Hengel~\cite{teney:eccv18} present a meta learning approach to
separate question answering from the information required for the
task, reasoning at test time over example questions and answers to
answer any given question. Two meta learning methods adapt a VQA model
without the need for retraining, and demonstrate the ability to
provide novel answers and support vision and language learning.
Rajani and Mooney~\cite{rajani:naacl18} developed an ensemble learning
approach, Stacking With Auxiliary Features, which combines the results
of multiple models using features of the problem as context.  The
approach considers four categories of auxiliary features, three of
which are inferred from image-question pairs while the fourth uses
model-specific explanations.

Research in cognitive systems indicates that reliable, efficient, and
explainable reasoning and learning can be achieved by reasoning with
domain knowledge and learning from experience. Early work by enabled
an agent to reason with first-order logic representations and
incrementally refined action operators~\cite{gil:icml94}. In such
methods, it is difficult to perform non-monotonic reasoning, or to
merge new, unreliable information with existing beliefs. Non-monotonic
logic formalisms have been developed to address these limitations,
e.g., Answer Set Prolog (ASP) has been used in cognitive
robotics~\cite{erdem:bookchap12} and other
applications~\cite{erdem:AIM16}. ASP has been combined with inductive
learning to monotonically learn causal laws~\cite{otero:ilp03}, and
methods have been developed to learn and revise domain knowledge
represented as ASP programs~\cite{balduccini:aaaisymp07,law:AIJ18}.
Cognitive architectures have also been developed to extract
information from perceptual inputs to revise domain knowledge
represented in first-order logic~\cite{laird:book12}, and to combine
logic and probabilistic representations to support reasoning and
learning in robotics~\cite{sarathy:TCDS16,zhang:TRO15}. However,
approaches based on classical first-order logic are not expressive
enough, e.g., modeling uncertainty by attaching probabilities to logic
statements is not always meaningful. Logic programming methods, on the
other hand, do not support one or more of the desired capabilities
such as efficient and incremental learning of knowledge, reasoning
efficiently with probabilistic components, or generalization as
described in this paper. These challenges can be addressed using
interactive task learning, a general knowledge acquisition framework
that uses labeled examples or reinforcement signals obtained from
observations, demonstrations, or human
instructions~\cite{chai:ijcai18,laird:IS17}. Sridharan and
Meadows~\cite{mohan:ACS18} developed such a framework to combine
non-monotonic logical reasoning with relational reinforcement learning
and inductive learning to learn action models to be used for reasoning
or learning in dynamic domains.  In the context of VQA, there has been
interesting work on reasoning with learned symbolic structure. For
instance, Yi et al.~\cite{yi:nips18} present a neural-symbolic VQA
system that uses deep networks to infer structural object-based scene
representation from images, and to generate a hierarchical (symbolic)
program of functional modules from the question. An executor then runs
the program on the representation to answer the question.  Such
approaches still do not (i) integrate reasoning and learning such that
they inform and guide each other; or (ii) use the rich domain-specific
commonsense knowledge that is available in any application domain.

In summary, deep networks represent the state of the art for VQA and
many other pattern recognition tasks. Recent surveys on VQA methods
indicate that despite considerable research, it is still difficult to
use these networks to support efficient learning, intuitive
explanations, or generalization to simulated and real-world
images~\cite{pandhre:TR17,shrestha:cvpr19}. Our architecture draws on
principles of cognitive systems to address these limitations. It
tightly couples deep networks with components for non-monotonic
logical reasoning with commonsense domain knowledge, and for learning
incrementally from samples over which the learned model makes errors.
This work builds on our proof of concept architecture that integrated
deep learning with commonsense inference for VQA~\cite{riley:hai18}.
It also builds on work in our research group on using commonsense
inference and learned state constraints to guide deep networks that
estimate object stability and occlusion in images~\cite{mota:rss19}.
In comparison with our prior work, we introduce a new component for
incrementally learning constraints governing domain states, expand
reasoning with commonsense knowledge to support planning and
diagnostics, explore the interplay between the architecture's
components, and discuss detailed experimental results.

\section{Architecture}
\label{sec:arch}
\begin{figure*}[tb]
  \centering
  \includegraphics[width=0.9\textwidth]{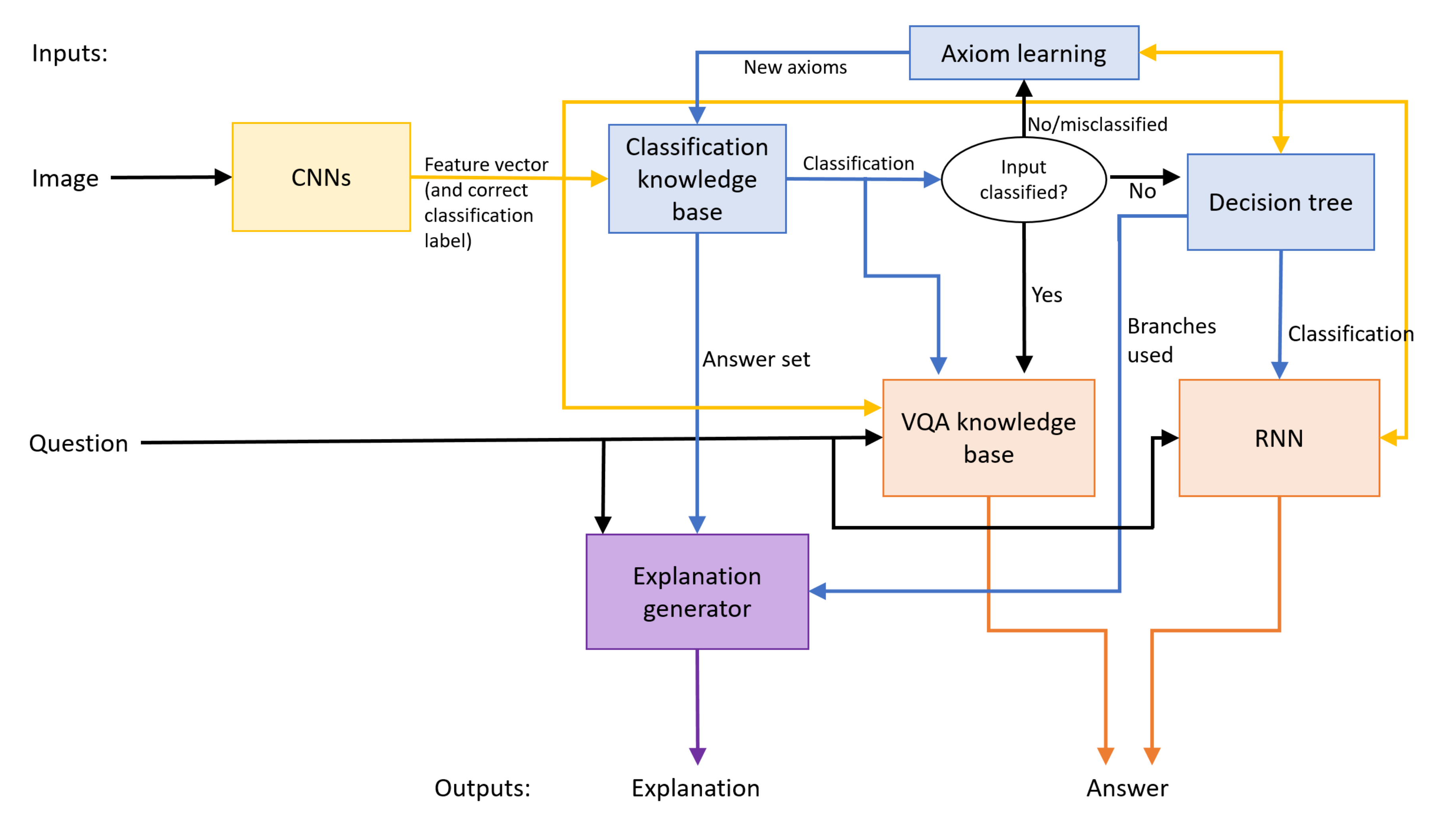}
  \vspace{-1em}
  \caption{Overview of the architecture that combines the principles
    of deep learning, non-monotonic logical reasoning, and
    decision-tree induction.}
  \label{fig:overall-arch}
\end{figure*}

Figure~\ref{fig:overall-arch} is an overview of our architecture that
provides answers to explanatory questions about images of scenes and
an underlying classification problem. The architecture seeks to
improve accuracy and reduce training effort, i.e., reduce training
time and the number of training samples, by embedding non-monotonic
logical reasoning and inductive learning in a deep network
architecture. We will later demonstrate how the architecture can be
adapted to address planning problems on a simulated robot---see
Section~\ref{sec:arch-plan}. The architecture may be viewed as having
four key components that are tightly coupled with each other.
\begin{enumerate}
\item A component comprising CNN-based feature extractors, which are
  trained and used to map any given image of a scene under
  consideration to a vector of image features.

\item A component that uses one of two methods to classify the feature
  vector. The first method uses non-monotonic reasoning with
  incomplete domain knowledge and the features to assign a class label
  and explain this decision. If the first method cannot classify the
  image, the second method trains and uses a decision tree to map the
  feature vector to a class label and explain the classification.

\item A component that answers explanatory questions. If non-monotonic
  logical reasoning is used for classification, it is also used to
  provide answers to these questions. If a decision tree is instead
  used for classification, an RNN is trained to map the decision
  tree's output, the image features, and the question, to the
  corresponding answer.

\item A component that uses the learned decision tree and the existing
  knowledge base to incrementally construct and validate constraints
  on the state of the domain. These constraints revise the existing
  knowledge that is used for subsequent reasoning.
\end{enumerate}
This architecture exploits the complementary strengths of deep
learning, non-monotonic logical reasoning, and incremental inductive
learning with decision trees. Reasoning with commonsense knowledge
guides learning, e.g., the RNN is trained on (and processes) input
data that cannot be processed using existing knowledge. The CNNs and
RNN can be replaced by other methods for extracting image features and
answering explanatory questions (respectively).  Also, although the
CNNs and RNN are trained in an initial phase in this paper, these
models can be revised over time if needed. We hypothesize that
embedding non-monotonic logical reasoning with commonsense knowledge
and the incremental updates of the decision tree, between the CNNs and
the RNN, makes the decisions more transparent, and makes learning more
time and sample efficient. Furthermore, the overall architecture and
methodology can be adapted to different domains. In this paper, we
will use the following two domains to illustrate and evaluate the
architecture's components and the methodology.

\begin{figure}[tb]
  \begin{center}
    \includegraphics[width=0.85\linewidth]{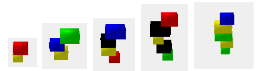}\hspace{0.05in}
    \vspace{-1em}
    \caption{Illustrative images of structures of blocks of different
      colors and sizes; these images were obtained from a
      physics-based simulator for the SS domain.}
    \label{fig:blocks-domain}
  \end{center}
  \vspace{-1em}
\end{figure}

\begin{figure}[tb]
  \begin{center}
    \includegraphics[height=0.17\columnwidth]{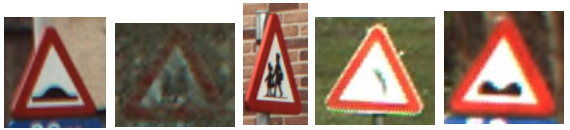}
    \caption{Illustrative images of traffic signs from the BelgiumTS
      dataset~\cite{timofte:ijcnn13}.}
    \label{fig:signs-domain}
    \vspace{-1em}
  \end{center}
\end{figure}

\begin{enumerate}
\item \textbf{Structure Stability (SS):} this domain has different
  structures, i.e., different arrangements of simulated blocks of
  different colors and sizes, on a tabletop---see
  Figure~\ref{fig:blocks-domain} for some examples. We generated
  $2500$ such images using a physics-based simulator. The relevant
  features of the domain include the number of blocks, whether the
  structure is on a lean, whether the structure has a narrow base, and
  whether any block is placed such that it is not well balanced on top
  of the block below. The objective in this domain is to classify
  structures as being stable or unstable, and to answer explanatory
  questions such as ``why is this structure unstable?''  and ``what
  should be done to make this structure stable?''.

\item \textbf{Traffic Sign (TS):} this domain focuses on recognizing
  traffic signs from images---see Figure~\ref{fig:signs-domain} for
  some examples. We used the BelgiumTS benchmark
  dataset~\cite{timofte:ijcnn13} with $\approx 7000$ real-world
  images (total) of $62$ different traffic signs. This domain's
  features include the primary symbol of the traffic sign, the
  secondary symbol, the shape of the sign, the main color in the
  middle, the border color, the sign's background image, and the
  presence or absence of a cross (e.g., some signs have a red or black
  cross across them to indicate the end of a zone, with the absence of
  the cross indicating the zone's beginning).
  The objective is to classify the traffic signs and answer
  explanatory questions such as ``what is the sign's message?''  and
  ``how should the driver respond to this sign?''.
\end{enumerate}
In addition to these two domains, Section~\ref{sec:arch-plan} will
introduce the \textbf{Robot Assistant (RA)} domain, a simulated domain
to demonstrate the use of our architecture for computing and executing
plans to achieve assigned goals. In the $RA$ domain, a simulated robot
reasons with existing knowledge to deliver messages to target people
in target locations, and to answer explanatory questions about the
plans and observed scenes.

The focus of our work is on understanding and using the interplay
between deep learning, commonsense reasoning, and incremental
learning, in the context of \emph{reliable and efficient scene
  understanding in any particular dynamic domain}. The benchmark VQA
datasets and the corresponding algorithms, on the other hand, focus on
generalizing across images from different scenarios in different
domains, making it difficult to support the reasoning and learning
capabilities of our architecture. We thus do not use these datasets or
algorithms in our evaluation.

\subsection{\bf Feature Extraction using CNNs}
\label{sec:arch-cnn}
\begin{figure*}
  \begin{center}
  \includegraphics[width=\linewidth]{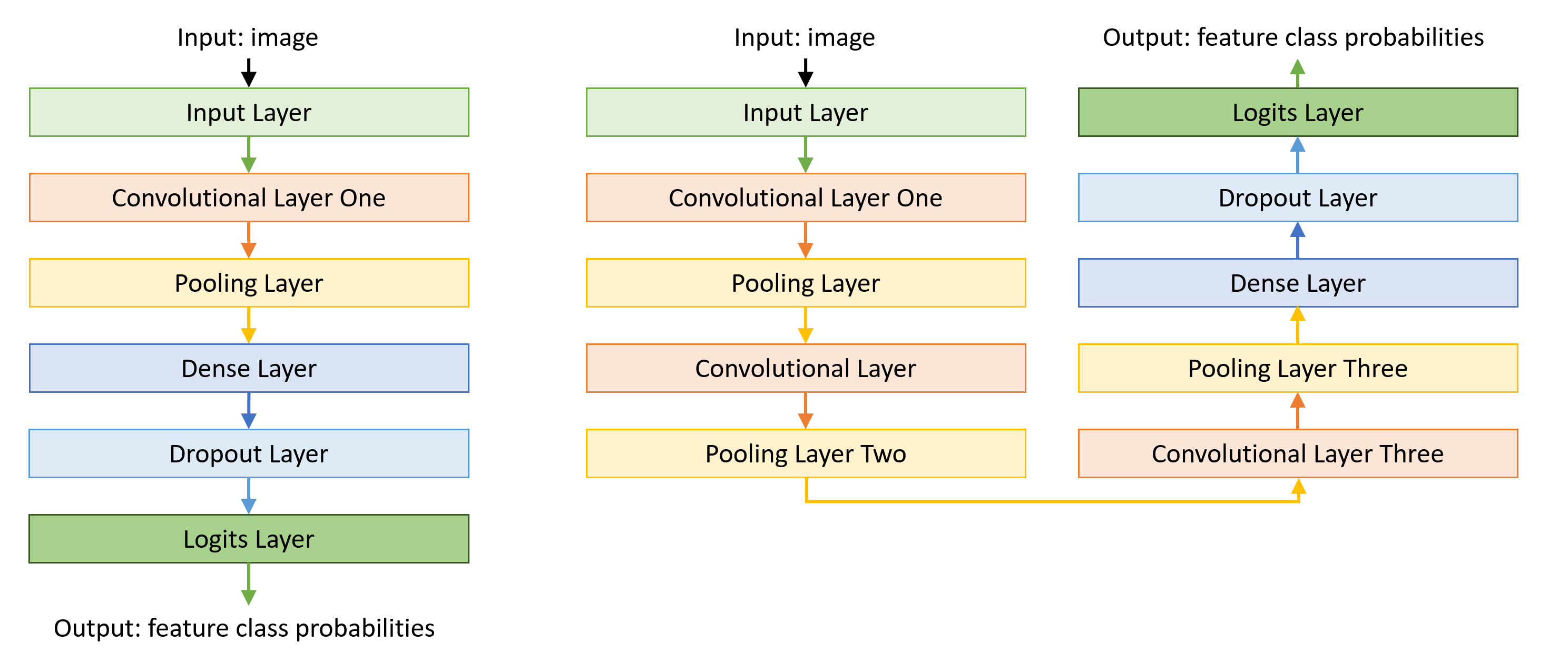}
  \vspace{-1em}
  \caption{Basic CNN model used for extracting each feature in our
    architecture. CNNs for individual features may end up with
    different numbers of convolutional and pooling layers.}
  \label{fig:cnn-example}
  \end{center}
  \vspace{-1em}
\end{figure*}

The first component of the architecture trains CNNs to map input
images to concise features representing the objects of interest in the
images. For the SS domain and TS domain, semi-automated annotation was
used to label the relevant features in images for training and
testing. The selection of these features for each domain was based on
domain expertise. In the SS domain, the features of interest are:
\begin{itemize}
\item Number of blocks in structure (number $\in [1, 5]$);
\item Whether the structure is on a lean (true, false);
\item Width of the base block (wide, narrow); and
\item Whether any block is displaced, i.e., placed such that it is not
  well balanced on top of the block below (true, false).
\end{itemize}
In the TS domain, the features of interest are:
\begin{itemize}
\item Primary symbol in the middle of the traffic sign; $39$ primary
  symbols such as $bumpy\_road$, $slippery\_road$, $stop$,
  $left\_turn$, and $speed\_limit$;
\item Secondary symbol in the traffic sign; $10$ secondary symbols
  such as $disabled$, $car$ and $fence$;
\item Shape of the sign; $circle$, $triangle$, $square$, $hexagon$,
  $rectangle$, $wide~rectangle$, $diamond$, or $inverted$ $triangle$;
\item Main color in the middle of the sign; $red$, $white$, or $blue$;
\item Border color at the edge of the sign; $red$, $white$, or $blue$;
\item Background image, e.g., some symbols are placed over a square or
  a triangle; and
\item Presence of a red or black cross across a sign to indicate a
  zone's end or invalidity; the sign without the cross indicates the
  zone's beginning or validity, e.g., a parking sign with a cross
  implies no parking.
\end{itemize}
To reduce the training data requirements and simplify the training of
CNNs, we (i) train a separate CNN for each feature to be extracted
from an image; and (ii) start with a basic model for each CNN and
incrementally make it more complex as needed. The number of CNNs is
thus equal to the number of features to be extracted from each image
for any given domain, and the CNN trained for each feature may be
different even within a particular domain. The basic CNN model we
begin with has an input layer, a convolutional layer, a pooling layer,
a dense layer, a dropout layer, and a logit layer, as seen on the left
of Figure~\ref{fig:cnn-example}. Additional convolutional and pooling
layers are added until the feature extraction accuracy converges or
exceeds a threshold (e.g., $\ge 90\%$). Our architecture also includes
the option of fine-tuning previously trained CNN models instead of
starting from scratch. The right side of Figure~\ref{fig:cnn-example}
shows a CNN model learned in our example domains, which has three
convolutional layers and pooling layers. We trained and validated
these CNNs in an initial phase, and used them for evaluation. Our code
for constructing these CNNs for features (in our example domains) is
in our repository~\cite{code-results}.

\subsection{\bf Classification using Non-monotonic Logical Reasoning
  or Decision Trees}
\label{sec:arch-classify}
The feature vector extracted from an image is used for decision
making. In the SS domain and TS domain, decisions take the form of
assigning a class label to each feature vector\footnote{Decision
  making can also include planning and diagnostics, as we will discuss
  later in Section~\ref{sec:arch-plan} for the RA domain.}. The second
component of our architecture performs this task using one of two
methods: (i) non-monotonic logical inference using ASP; or (ii) a
classifier based on a learned decision tree.  We describe these two
methods below.

\medskip
\noindent
\textbf{ASP-based Inference with Commonsense Knowledge:} ASP is a
declarative programming paradigm that can be used to represent and
reason with incomplete commonsense domain knowledge. ASP is based on
stable model semantics, and supports \emph{default negation} and
\emph{epistemic disjunction}. For instance, unlike ``$\lnot a$'',
which implies that \emph{a is believed to be false}, ``$not~a$'' only
implies \emph{a is not believed to be true}. Also, unlike
``$p~\lor\,\,\lnot p$'' in propositional logic, ``$p~or~\lnot p$'' is
not tautological.  Each literal can thus be true, false or unknown,
and the agent reasoning with domain knowledge does not believe
anything that it is not forced to believe. ASP can represent recursive
definitions, defaults, causal relations, special forms of
self-reference, and language constructs that occur frequently in
non-mathematical domains, and are difficult to express in classical
logic formalisms~\cite{baral:book03,gelfond:aibook14}. Unlike
classical first-order logic, ASP supports non-monotonic logical
reasoning, i.e., it can revise previously held conclusions or
equivalently reduce the set of inferred consequences, based on new
evidence---this ability helps the agent recover from any errors made
by reasoning with incomplete knowledge. ASP and other paradigms that
reason with domain knowledge are often criticized for requiring
considerable (if not complete) prior knowledge and manual supervision,
and for being unwieldy in large, complex domains. However, modern ASP
solvers support efficient reasoning in large knowledge bases with
incomplete knowledge, and are used by an international research
community for cognitive robotics~\cite{erdem:bookchap12,zhang:TRO15}
and other applications~\cite{erdem:AIM16}. For instance, recent work
has demonstrated that ASP-based non-monotonic logical reasoning can be
combined with: (i) probabilistic reasoning for reliable and efficient
planning and diagnostics~\cite{mohan:JAIR19}; and (ii) relational
reinforcement learning and active learning methods for interactively
learning or revising commonsense domain knowledge based on input from
sensors and humans~\cite{mohan:ACS18}.

A domain's description (i.e., the knowledge base) in ASP comprises a
\emph{system description} $\mathcal{D}$ and a \emph{history}
$\mathcal{H}$. System description $\mathcal{D}$ comprises a
\emph{sorted signature} $\Sigma$ and axioms. The signature $\Sigma$
comprises \emph{basic sorts}, \emph{statics}, i.e., domain attributes
whose values do not change over time, \emph{fluents}, i.e., domain
attributes whose values can change over time, and \emph{actions}. The
domain's fluents can be \emph{basic}, i.e., those that obey the laws
of inertia and are changed directly by actions, or \emph{defined},
i.e., those that do not obey the laws of inertia and are defined by
other attributes. For instance, in the SS domain, $\Sigma$ includes
basic sorts such as $structure$, $color$, $size$, and $attribute$; the
basic sorts of the TS domain include $main\_color$, $other\_color$,
$main\_symbol$, $other\_symbol$, $shape$, $cross$ etc. The sort $step$
is also in $\Sigma$ to support temporal reasoning over time steps. The
statics and fluents in the SS domain include: 
\vspace{-1.5em}
\begin{center}
  \begin{align}
    &num\_blocks(structure, num), ~block\_color(block, color),~block\_size(block, size) \\ \nonumber
    &block\_displaced(structure), ~stable(structure)
  \end{align}
\end{center}
which correspond to the image features extracted in the domain, and
are described in terms of their arguments' sorts. In a similar manner,
statics and fluents of the TS domain include: 
\vspace{-1.5em}
\begin{center}
  \begin{align}
    &primary\_symbol(sign, main\_symbol), ~primary\_color(sign, main\_color)\\\nonumber
    &secondary\_symbol(sign, other\_symbol), ~secondary\_color(sign, other\_color)\\ \nonumber
    &sign\_shape(shape), ~background\_image(image)
  \end{align}
\end{center}
In both domains, signature $\Sigma$ includes a predicate
$holds(fluent, step)$, which implies that a particular fluent holds
true at a particular time step. As stated above, $\Sigma$ for a
dynamic domain typically includes actions that cause state
transitions, but this capability is not needed to answer explanatory
questions about specific scenes and the underlying classification
problem in our (SS, TS) domains. For ease of explanation, we thus
temporarily disregard the modeling of actions, and their preconditions
and effects. We will revisit actions in Section~\ref{sec:arch-plan}
when we consider planning tasks in the RA domain.

The axioms of $\mathcal{D}$ govern domain dynamics; in our example
domains (SS, TS), they govern the belief about domain states. The
axioms of the SS domain include statements such as:
\begin{subequations}
  \label{eqn:axioms-ss}
  \begin{align}
    &\lnot stable(S) ~~\leftarrow~~block\_displaced(S)\\
    &stable(S) ~~\leftarrow~~ num\_blocks(S, 2),~structure\_type(S, lean)
  \end{align}
\end{subequations}
where Statement~\ref{eqn:axioms-ss}(a) says that any structure with a
block that is displaced significantly is unstable, and
Statement~\ref{eqn:axioms-ss}(b) says that any pair of blocks without
a significant lean is stable.

\noindent
Axioms of the TS domain include statements such as:
\begin{subequations}
  \label{eqn:axioms-ts}
  \begin{align} 
    \nonumber sign\_type(TS, no\_parking) \leftarrow~ &primary\_color(TS,
    blue),~primary\_symbol(TS, blank), \\ &cross(TS),~shape(TS, circle) \\\nonumber
    sign\_type(TS, stop) \leftarrow &primary\_color(TS,
    red),~primary\_symbol(TS, stoptext), \\
    & shape(TS, octagon)
  \end{align}
\end{subequations}
where Statement~\ref{eqn:axioms-ts}(a) implies that a traffic sign
that is blue, blank, circular, and has a cross across it, is a no
parking sign. Statement~\ref{eqn:axioms-ts}(b) implies that a traffic
sign that is red, octagon-shaped, and contains the text ``stop'', is a
stop sign.

The history $\mathcal{H}$ of a dynamic domain is usually a record of
fluents observed to be true or false at a particular time step, i.e.,
$obs(fluent, boolean, step)$, and the successful execution of an
action at a particular time step, i.e., $hpd(action, step)$; for more
details, see~\cite{gelfond:aibook14}. The domain knowledge in many
domains often includes default statements that are true in all but a
few exceptional circumstances. For example, we may know in the SS
domain that ``structures with two blocks of the same size are usually
stable''. To encode such knowledge, we use our recent work that
expanded the notion of history to represent and reason with defaults
describing the values of fluents in the initial
state~\cite{mohan:JAIR19}.

Key tasks of an agent equipped with a system description $\mathcal{D}$
and history $\mathcal{H}$ include reasoning with this knowledge for
inference, planning and diagnostics. In our architecture, these tasks
are accomplished by translating the domain representation to a program
$\Pi(\mathcal{D}, \mathcal{H})$ in CR-Prolog, a variant of ASP that
incorporates consistency restoring (CR)
rules~\cite{balduccini:aaaisymp03}. In this paper, we use the terms
``ASP'' and ``CR-Prolog'' interchangeably. The program $\Pi$ includes
the signature and axioms of $\mathcal{D}$, inertia axioms, reality
checks, closed world assumptions for defined fluents and actions, and
observations, actions, and defaults from $\mathcal{H}$.  In addition,
features extracted from an input image (to be processed) are encoded
as the initial state of the domain in $\Pi$. Each \emph{answer set} of
$\Pi(\mathcal{D}, \mathcal{H})$ represents the set of beliefs of an
agent associated with this program.  Algorithms for computing
entailment, and for planning and diagnostics, reduce these tasks to
computing answer sets of CR-Prolog programs. We compute answer sets of
CR-Prolog programs using the SPARC system~\cite{balai:lpnmr13}.  The
CR-Prolog programs for our example domains are in our open-source
software repository~\cite{code-results}. For the classification task
in our example domains, the relevant literals in the answer set
provide the class label and an explanatory description of the assigned
label (see Section~\ref{sec:arch-explain}); we will consider the
planning task in Section~\ref{sec:arch-plan}. The accuracy of the
inferences drawn from the encoded knowledge depends on the accuracy
and extent of the knowledge encoded, but encoding comprehensive domain
knowledge is difficult. The decision of what and how much knowledge to
encode is made by the designer.

\medskip
\noindent
\textbf{Decision Tree Classifier:} If ASP-based inference cannot
classify the feature vector extracted from an image, the feature
vector is mapped to a class label using a decision tree classifier
learned from labeled training examples. In a decision tree classifier,
each node is associated with a question about the value of a
particular feature, with the child nodes representing the different
answers to the question, i.e., the possible values of the feature.
Each node is also associated with samples that satisfy the
corresponding values of the features along the path from the root node
to this node. We use a standard implementation of a decision tree
classifier~\cite{duda:book00}. This implementation uses the Gini
measure to compute information gain (equivalently, the reduction in
entropy) that would be achieved by splitting an existing node based on
each feature that has not already been used to create a split in the
tree. Among the features that provide a significant information gain,
the feature that provides the maximum information gain is selected to
split the node. If none of the features would result in any
significant information gain, this node becomes a leaf node with a
class label that matches a majority of the samples at the node.

\begin{figure*}
  \includegraphics[width=\linewidth]{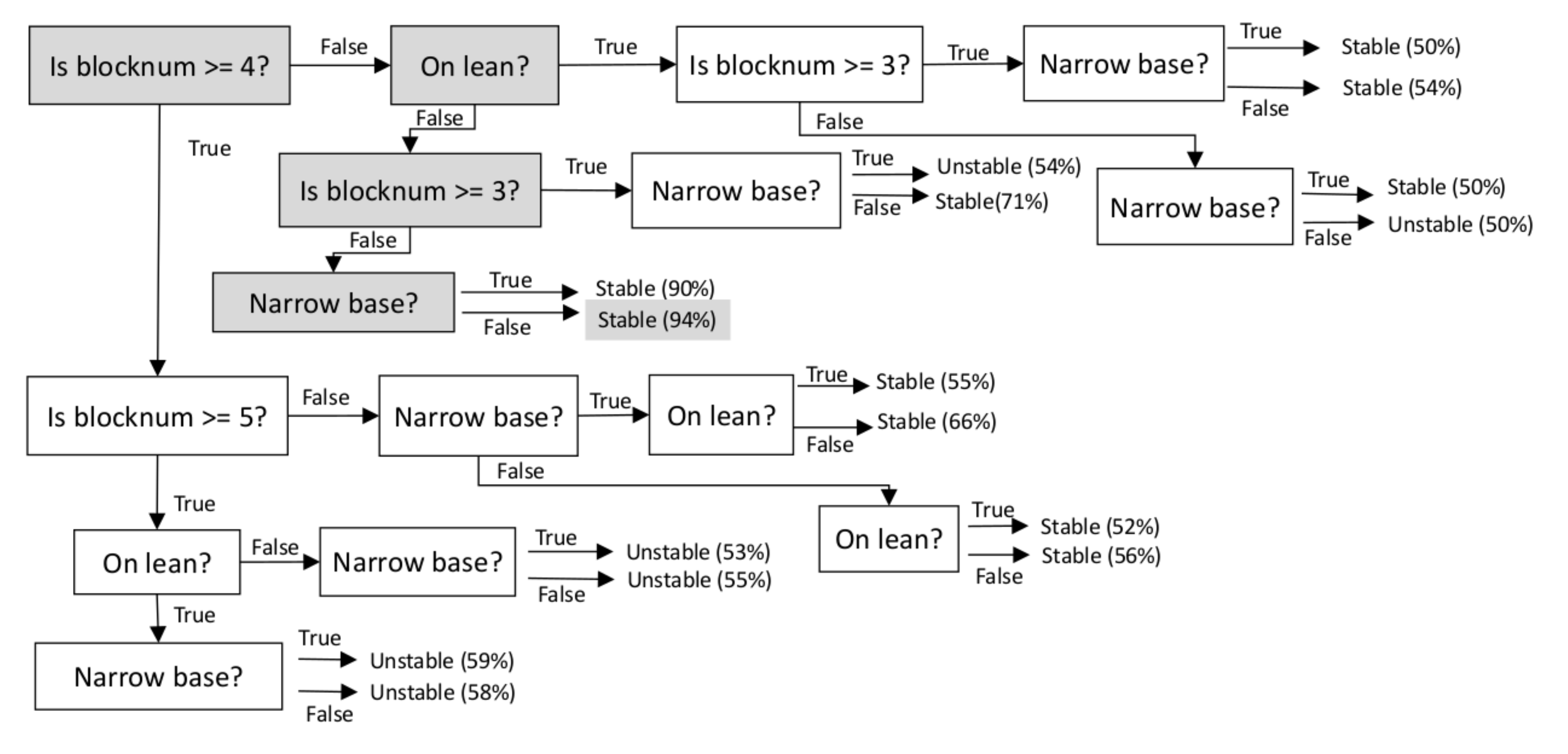}
  \vspace{-1.5em}
  \caption{Example of part of a decision tree constructed from labeled
    samples and used for classification in the SS domain. The nodes
    used to classify a particular example are highlighted. Each leaf
    shows a class label and indicates the proportion of the labeled
    examples (at the leaf) that correspond to this label.}
  \label{fig:decisiontree-example}
\end{figure*}

The decision tree's search space is quite specific since it only
considers samples that could not be classified by ASP-based reasoning.
The decision tree does not need to generalize as much as it would have
to if it had to process every training (or test) sample in the
dataset. Also, although overfitting is much less likely, we still use
pruning to minimize the effects of overfitting.
Figure~\ref{fig:decisiontree-example} shows part of a learned decision
tree classifier; specific nodes used to classify a particular example
are highlighted to indicate that $94\%$ of the observed examples of
structures that have fewer than three blocks, do not have a
significant lean, and do not have a narrow base, correspond to stable
structures. These ``active'' nodes along any path in the decision tree
that is used to classify an example can be used to explain the
classification outcome in terms of the values of particular features
that were used to arrive at the class label assigned to a specific
image under consideration.

\subsection{\bf Answering Explanatory Questions}
\label{sec:arch-explain}
The third component of the architecture provides two methods for
answering explanatory questions. The available inputs are the (i)
question; (ii) vector of features extracted from the image under
consideration; and (iii) classification output. The human designer
also provides pre-determined templates for questions and their
answers. In our case, we use a controlled vocabulary, templates based
on language models and parts of speech for sentences, and existing
software for natural language processing. Any given question is
transcribed using the controlled vocabulary, parsed (e.g., to obtain
parts of speech), and matched with the templates to obtain a
relational representation. Recall that questions in the SS domain are
of the form: ``is this structure stable/unstable?'' and `` what is
making this structure stable/unstable?''. These questions can be
translated into relational statements such as $stable(S)$ or $\lnot
stable(S)$ and used as a question, or as the desired consequence,
during inference or in a search process. In a similar manner,
questions in the TS domain such as: ``what sign is this?'' and ``what
is the sign's message?''  can be translated into $sign\_type(S, sign)$
and used for subsequent processing.

The first method for answering explanatory questions is based on the
understanding that if the feature vector extracted from the image is
processed successfully using ASP-based reasoning, it is also possible
to reason with the existing knowledge to answer explanatory questions
about the scene. To support such question answering, we need to revise
the signature $\Sigma$ in the system description $\mathcal{D}$ of the
domain. For instance, we add sorts such as $query\_type$,
$answer\_type$, and $query$ to encode different types of queries and
answers. We also introduce suitable relations to represent questions,
answers to these questions, and more abstract attributes, e.g., of
structures of blocks, traffic signs etc.

In addition to the signature, we also augment the axioms in
$\mathcal{D}$ to support reasoning with more abstract attributes, and
to help construct answers to questions. For instance, we can include
an axiom such as:
\begin{align} 
  \nonumber 
  many\_blocks(S)~\leftarrow~ &unstable(S),~\lnot base(S,
  narrow),\\ &\lnot struc\_type(S, lean),~\lnot
  block\_displaced(S)
\end{align}
which implies that if a structure (of blocks) is not on a narrow base,
does not have a significant lean, and does not have blocks
significantly displaced, any instability in the structure implies (and
is potentially because) there are too many blocks in the structure. 
Once the ASP program $\Pi(\mathcal{D}, \mathcal{H})$ has been revised
as described above, we can compute answer set(s) of this program to
obtain the beliefs of the agent associated with this program. For any
given question, the answer set(s) are parsed based on the known
controlled vocabulary and templates (for questions and answers) to
extract relevant literals---these literals are included in the
corresponding templates to construct answers to explanatory questions.
These answers can also be converted to speech using existing software.

\begin{figure}
  \begin{center}
    \includegraphics[width=0.55\columnwidth]{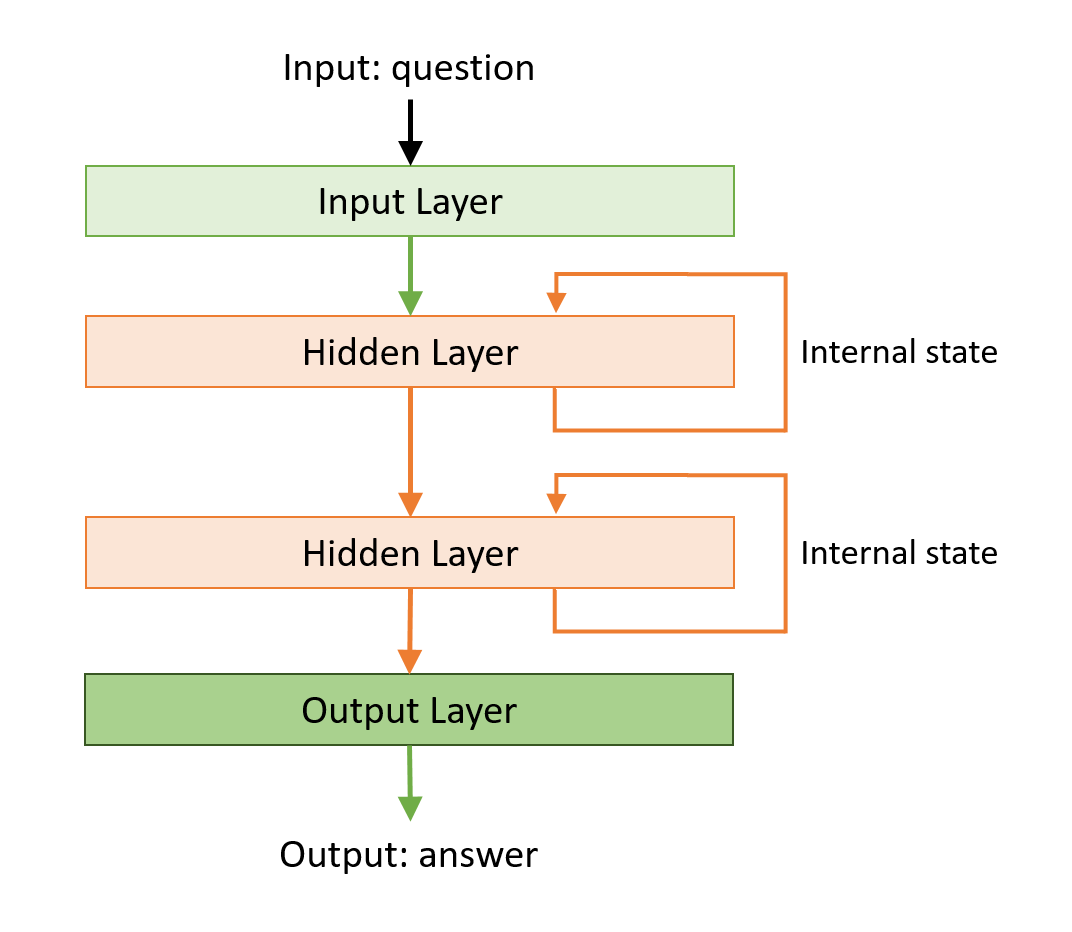}
    \vspace{-2em}
    \caption{Example of the basic RNN used to construct explanations.
      The RNN learned for the example domains has $26-30$ hidden
      layers.}
    \label{fig:rnn-example}
  \end{center}
\end{figure}
 
The second method for answering explanatory questions is invoked if
the decision tree is used to process (i.e., classify in the context of
the SS domain and TS domain) the vector of image features.  The
inability to classify the feature vector through ASP-based reasoning
is taken to imply that the encoded domain knowledge is insufficient to
answer explanatory questions about the scene. In this case, an LSTM
network-based RNN is trained and used to answer the explanatory
questions. The inputs are the feature vector, classification output,
and a vector representing the transcribed and parsed query. The output
(provided during training) is in the form of answers in the
predetermined templates. Similar to the approach used in
Section~\ref{sec:arch-classify}, the RNN is built incrementally during
training. We begin with one or two hidden layer(s), as shown in
Figure~\ref{fig:rnn-example}, and add layers as long as it results in
a significant increase in the accuracy.  We also include the option of
adding a stack of LSTMs if adding individual layers does not improve
accuracy significantly. In our example domains, the RNN constructed to
answer explanatory questions had as many as $26-30$ hidden layers and
used a softmax function to provide one of about $50$ potential answer
types. An example of the code used to train the RNN is available in
our repository~\cite{code-results}.

\subsection{\bf Learning State Constraints}
\label{sec:arch-learn-constraint}
The components of the architecture described so far support reasoning
with commonsense knowledge, learned decision trees, and deep networks,
to answer explanatory questions about the scene and an underlying
classification problem. In many practical domains, the available
knowledge is incomplete, the number of labeled examples is small, or
the encoded knowledge changes over time. The decisions made by the
architecture can thus be incorrect or sub-optimal, e.g., a traffic
sign can be misclassified or an ambiguous answer may be provided to an
explanatory question. The fourth component of our architecture seeks
to address this problem by supporting incremental learning of domain
knowledge. Our approach is inspired by the inductive learning methods
mentioned in Section~\ref{sec:relwork}, e.g., work in our research
group on using relational reinforcement learning and decision tree
induction to learn domain axioms~\cite{mohan:ACS18}. The work
described in this paper uses decision tree induction to learn
constraints governing domain states. The methodology used in this
component, in the context of VQA, is as follows:
\begin{enumerate}
\item Identify training examples that are not classified, or are
  classified incorrectly, using the existing knowledge. Recall that
  this step is accomplished by the component described in
  Section~\ref{sec:arch-classify}, which processes each training
  example using the existing knowledge encoded in the CR-Prolog
  program, in an attempt to assign a class label to the example.

\item Train a decision tree using the examples identified in Step-1
  above. Recall that this step is also accomplished by the component
  described in Section~\ref{sec:arch-classify}.

\item Identify paths in the decision tree (from root to leaf) such
  that (i) there are a sufficient number of examples at the leaf,
  e.g., $10\%$ of the training examples; and (ii) all the examples at
  the leaf have the same class label. Since the nodes correspond to
  checks on the values of domain features, the paths will correspond
  to combinations of partial state descriptions and class labels that
  have good support among the labeled training examples. Each such
  path is translated into a candidate axiom. For instance, the
  following are two axioms identified by this approach in the SS
  domain:
  \begin{subequations}
    \label{eqn:axiom-learn}
    \begin{align}
      &\lnot stable(S) ~:-~num\_blocks(S, 3),~ base(S, wide),~struc\_type(S, lean)\\
      &\lnot stable(S) ~:-~num\_blocks(S, 3),~ base(S, narrow),~struc\_type(S, lean)
    \end{align}
  \end{subequations}

\item Generalize candidate axioms if possible. For instance, if one
  candidate axiom is a over-specification of another existing axiom,
  the over-specified version is removed. In the context of the axioms
  in Statement~\ref{eqn:axiom-learn}(a-b), the second literal
  represents redundant information, i.e., if a structure with three
  blocks has a significant lean, it is unstable irrespective of
  whether the base of the structure is narrow or wide. Generalizing
  over these two axioms results in the following candidate axiom:
  \begin{align}
    &\lnot stable(S) ~:-~num\_blocks(S, 3),~struc\_type(S, lean)
  \end{align}
  which only includes the literals that encode the essential
  information.

\item Validate candidate axioms one at a time. To do so, the candidate
  axiom is added to the CR-Prolog program encoding the domain
  knowledge. A sufficient number of training examples (e.g., $10\%$ of
  the dataset, as before) relevant to this axiom, i.e., the domain
  features encoded by the examples should satisfy the body of the
  axiom, are drawn randomly from the training dataset. If processing
  these selected examples with the updated CR-Prolog program results
  in misclassifications, the candidate axiom is removed from further
  consideration.

\item Apply sanity checks to the validated axioms. The validated
  axioms and existing axioms are checked to remove over-specifications
  and retain the most generic version of any axiom. Axioms that pass
  these sanity checks are added to the CR-Prolog program and used for
  subsequent reasoning.
\end{enumerate}
Section~\ref{sec:expresults-axiom-vqa} examines the effect of such
learned constraints on classification and VQA performance.

\subsection{\bf Planning with Domain Knowledge}
\label{sec:arch-plan}
\begin{figure*}
  \begin{center}
    \includegraphics[width=0.55\textwidth]{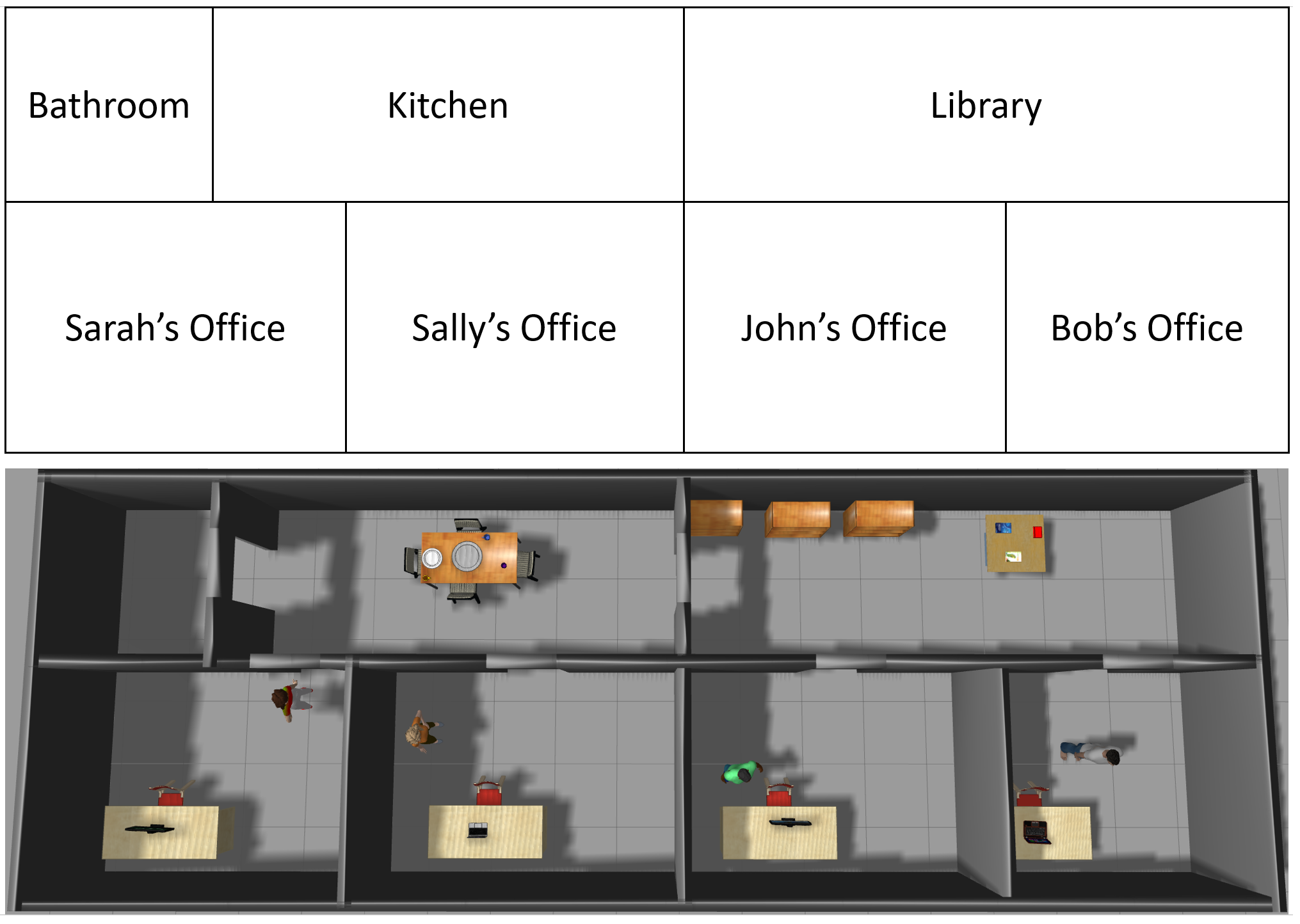}
    \vspace{-0.75em}
    \caption{Block diagram and a simulated scenario in the RA domain
      in which the robot has to deliver messages to people in target
      locations.}
    \vspace{-1.25em}
    \label{fig:ra-domain}
  \end{center}
\end{figure*}

The description of the architecture's components has so far focused on
classification and VQA, and reasoning has been limited to inference
with knowledge. However, the architecture is also applicable to
planning (and diagnostics) problems. Consider the \textbf{RA domain}
in which a simulated robot has to navigate and deliver messages to
particular people in particular places, and to answer explanatory
questions, i.e., the domain includes aspects of planning and VQA.
Figure~\ref{fig:ra-domain} depicts this domain and a simulated
scenario in it; semantic labels of the offices and rooms are shown in
the upper half.

A robot planning and executing actions in the real world has to
account for the uncertainty in sensing and actuation. In other work,
we addressed this issue by coupling ASP-based coarse-resolution
planning with probabilistic fine-resolution planning and
execution~\cite{mohan:JAIR19}. In this paper, we temporarily abstract
away such probabilistic models of uncertainty to thoroughly explore
the interplay between reasoning and learning, including the effect of
added noise in sensing and actuation (in simulation).

To support planning, the signature $\Sigma$ of system description
$\mathcal{D}$ has basic sorts such as: $place$, $robot$, $person$,
$object$, $entity$, $status$, and $step$, which are arranged
hierarchically, e.g., $robot$ and $person$ are subsorts of $agent$,
and $agent$ and $object$ are subsorts of $entity$. $\Sigma$ also
includes ground instances of sorts, e.g., $of\!fice$, $workshop$,
$kitchen$, and $library$ are instances of $place$, and $Sarah$, $Bob$,
$John$, and $Sally$ are instances of $person$. As before, domain
attributes and actions are described in terms of the sorts of their
arguments. The fluents include $loc(agent, place)$, which describes
the location of the robot and people in the domain, and
$message\_status(message\_id, person, status)$, which denotes whether
a particular message has been delivered (or remains undelivered) to a
particular person. Static attributes include relations such as
$next\_to(place, place)$ and $work\_place(person, place)$ to encode
the arrangement of places and the work location of people
(respectively) in the domain.  Actions of the domain include:
\begin{align}
  &move(robot, place) \\ \nonumber
  &deliver(robot, message\_id, person)
\end{align}
which move the robot to a particular place, and cause a robot to
deliver a particular message to a particular person (respectively).
For ease of explanation, we assume that the locations of people are
defined fluents whose values are determined by external sensors, and
that the locations of objects are static attributes; as a result, we
do not consider actions that change the value of these attributes.
The signature $\Sigma$ also includes (as before) the relation
$holds(fluent, step)$ to imply that a particular fluent is true at a
particular time step.

Axioms of the RA domain capture the domain's dynamics. These axioms
include causal laws, state constraints and executability conditions.
For example:
\begin{subequations}
  \label{eqn:axioms-ra}
  \begin{align}
    &move(rob_1, L)~~\mathbf{causes}~~loc(rob_1, L)\\
    &deliver(rob_1, ID, P)~~\mathbf{causes}~~message\_status(ID, P, delivered) \\
    &loc(P, L) ~~\mathbf{if}~~ work\_place(P, L),~not~\lnot loc(P, L)\\
    &\lnot loc(T, L_2)~~\mathbf{if}~~ loc(T, L_1),~L_1 \ne L_2 \\
    &\mathbf{impossible}~~deliver(rob_1, ID, P) ~~\mathbf{if}~loc(rob_1, L_1),~loc(P, L_2),~L_1 \ne L_2 \\
    &\mathbf{impossible}~~move(rob_1, L) ~~\mathbf{if}~~ loc(rob_1, L)
  \end{align}
\end{subequations}
where Statement~\ref{eqn:axioms-ra}(a) states that executing a move
action causes the robot's location to be the target place;
Statement~\ref{eqn:axioms-ra}(b) states that executing a deliver
action causes the message to be delivered to the desired person;
Statement~\ref{eqn:axioms-ra}(c) is a constraint stating that unless
told otherwise the robot expects (by default) a person to be in
her/his place of work; Statement~\ref{eqn:axioms-ra}(d) is a
constraint stating that any thing can be in one place at at time;
Statement~\ref{eqn:axioms-ra}(e) implies that a robot cannot deliver a
message to an intended recipient if the robot and the person are not
in the same place; and Statement~\ref{eqn:axioms-ra}(f) states that a
robot cannot move to a location if it is already there.
 
As described in Section~\ref{sec:arch-classify}, the domain history is
a record of observations (of fluents), the execution of actions, and
the values of fluents in the initial state. Also, planning (similar to
inference) is reduced to computing answer set(s) of the program
$\Pi(\mathcal{D}, \mathcal{H})$ after including some helper axioms for
computing a minimal sequence of actions; for examples, please
see~\cite{gelfond:aibook14,mohan:JAIR19}. If the robot's knowledge of
the domain is incomplete or incorrect, the computed plans may be
suboptimal or incorrect. The approach described in
Section~\ref{sec:arch-learn-constraint} can then be used to learn the
missing constraints; we will explore the interplay between learning
and planning in Section~\ref{sec:expresults-axiom-plan}.

\section{Experimental Setup and Results}
\label{sec:expresults}
In this section, we describe the results of experimentally evaluating
the following hypotheses about the capabilities of our architecture:
\begin{itemize}
\item \underline{\textbf{H1:}} for the underlying classification
  problem, our architecture outperforms an architecture based on just
  deep networks for small training datasets, and provides comparable
  performance as the size of the dataset increases;
\item \underline{\textbf{H2:}} in the context of answering explanatory
  questions, our architecture provides significantly better
  performance in comparison with an architecture based on deep
  networks;
\item \underline{\textbf{H3:}} our architecture supports reliable and
  incremental learning of state constraints, which improves the
  ability to answer explanatory questions; and
\item \underline{\textbf{H4:}} our architecture can be adapted to
  planning tasks, with the incremental learning capability improving
  the ability to compute minimal plans.
\end{itemize}
These hypotheses were evaluated in the context of the domains (SS, TS
and RA) introduced in Section~\ref{sec:arch}. Specifically, hypotheses
$H1$, $H2$ and $H3$ are evaluated in the SS domain and TS domain in
the context of VQA. As stated in Section~\ref{sec:introduction}, VQA
is used in this paper only as an instance of a complex task that
requires explainable reasoning and learning. We are primarily
interested in exploring the interplay between reasoning with
commonsense domain knowledge, incremental learning, and deep learning,
in any given domain in which large labeled datasets are not readily
available.  State of the art VQA algorithms, on the other hand, focus
instead on generalizing across different domains, using benchmark
datasets of several thousand images. Given the difference in
objectives between over work and the existing work on VQA, we thus do
not compare with state of the art algorithms, and do not use the
benchmark VQA datasets. Furthermore, we evaluated hypothesis $H4$ in
the RA domain in which the robot's goal was to deliver messages to
appropriate people and answer explanatory questions about this
process.

We begin by describing some execution traces in
Section~\ref{sec:expresults-trace} to illustrate the working of our
architecture. This is followed by
Sections~\ref{sec:expresults-classify-vqa}-~\ref{sec:expresults-axiom-plan},
which describe the results of experimentally evaluating the
classification, VQA, axiom learning, and planning capabilities, i.e.,
hypotheses $H1$-$H4$. We use accuracy (precision) as the primary
performance measure.  Classification accuracy was measured by
comparing the assigned labels with the ground truth values, and
question answering accuracy was evaluated heuristically by computing
whether the answer mentions all image attributes relevant to the
question posed. This relevance was established by a human expert, the
lead author of this paper. Unless stated otherwise, we used two-thirds
of the available data to train the deep networks and other
computational models, using the remaining one-third of the data for
testing. For each image, we randomly chose from the set of suitable
questions for training the computational models. We repeated this
process multiple times and report the average of the results obtained
in these trials. Finally, for planning, accuracy was measured as the
ability to compute minimal and correct plans for the assigned goals.

\subsection{\bf Execution Traces}
\label{sec:expresults-trace}
The following execution traces illustrate our architecture's ability
to reason with commonsense knowledge and learned models to provide
intuitive answers for explanatory questions.


\noindent
\begin{execexample}\label{ex:exec-example1}[Question Answering, SS
  domain]
  {\rm Consider a scenario in the SS Domain in which the input (test)
    image is the one on the extreme right in
    Figure~\ref{fig:blocks-domain}.
    \begin{itemize}
    \item First \textbf{classification-related question} posed:
      ``\emph{is this structure unstable}?'' \ \\
      The architecture's \textbf{answer}: ``\emph{no}''.

    \item The \textbf{explanatory question} posed: ``\emph{what is
        making this structure stable}?'' \ \\
      The architecture's \textbf{answer}: ``\emph{the structure has
        five blocks and a narrow base, it is standing straight, and
        there is no significant lean}''.

    \item This answer was based on the following features extracted by
      CNNs from the image: (i) five blocks; (ii) narrow base; (iii)
      standing straight; and (iv) no significant lean, i.e, all blocks
      in place.

    \item The extracted features wew converted to literals. ASP-based
      inference provided an answer about the stability of the
      arrangement of objects in the scenario. Relevant literals in the
      corresponding answer set were then inserted into a suitable
      template to provide the answers described above.
      
    \item Since the example was processed successfully using ASP-based
      inference, it was not processed using the decision tree (for
      classification) or the RNN (for answering the explanatory
      question).
    \end{itemize}
  }
\end{execexample}

\begin{execexample}\label{ex:exec-example2}[Question Answering, TS
  domain]
  {\rm Consider a scenario in the TS Domain with the input (test)
    image is the one on the extreme right in
    Figure~\ref{fig:signs-domain}.
    \begin{itemize}
    \item The \textbf{classification question} posed was: ``\emph{what
        is the sign's message}?''\ \\
      The architecture's \textbf{answer}: ``\emph{uneven surfaces ahead}''.

    \item When asked to explain this answer (``\emph{Please explain
        this answer}''), the architecture identified that the CNNs
      extracted the following features of the sign in the image: (i)
      it is triangle-shaped; (ii) main color is white and other (i.e.,
      border) color is red; (iii) it has no background image; (iv) it
      has a bumpy-road symbol and no secondary symbol; and (v) it has
      no cross.

    \item These features were converted to literals and used in
      ASP-based inference based on existing knowledge in the TS
      domain. ASP-based inference is unable to provide an answer,
      i.e., unable to classify the sign.

    \item The extracted features were processed using the trained
      decision tree, which only used the colors in the sign to assign
      the class label. The main (or border) color is normally
      insufficient to accurately classify signs. However, recall that
      the decision tree is trained to classify signs that cannot be
      classified by reasoning with existing knowledge.

    \item The decision tree output, image feature vector, and input
      question, were processed by the previously trained RNN to
      provide the answer type and the particular answer described
      above.
    \end{itemize}
  }
\end{execexample}
\noindent
These (and other such) execution traces illustrate the working of our
architecture, especially that:
\begin{itemize}
\item The architecture takes advantage of (and perform non-monotonic
  logical inference with) the existing commonsense domain knowledge to
  reliably and efficiently address the decision-making problem
  (classification in the examples above) when possible. In such cases,
  it is also able to answer explanatory questions about the
  classification decision and the underlying scene.

\item When the desired decision cannot be made using non-monotonic
  logical inference with domain knowledge, the architecture smoothly
  transitions to training and using a decision-tree to make and
  explain the classification decision. In such cases, the architecture
  also learns and uses an RNN to answer explanatory questions about
  the scene.
\end{itemize}

\subsection{\bf Experimental Results: Classification + VQA}
\label{sec:expresults-classify-vqa}
\begin{figure*}[tbh]
  \centering
  \includegraphics[width=0.7\linewidth]{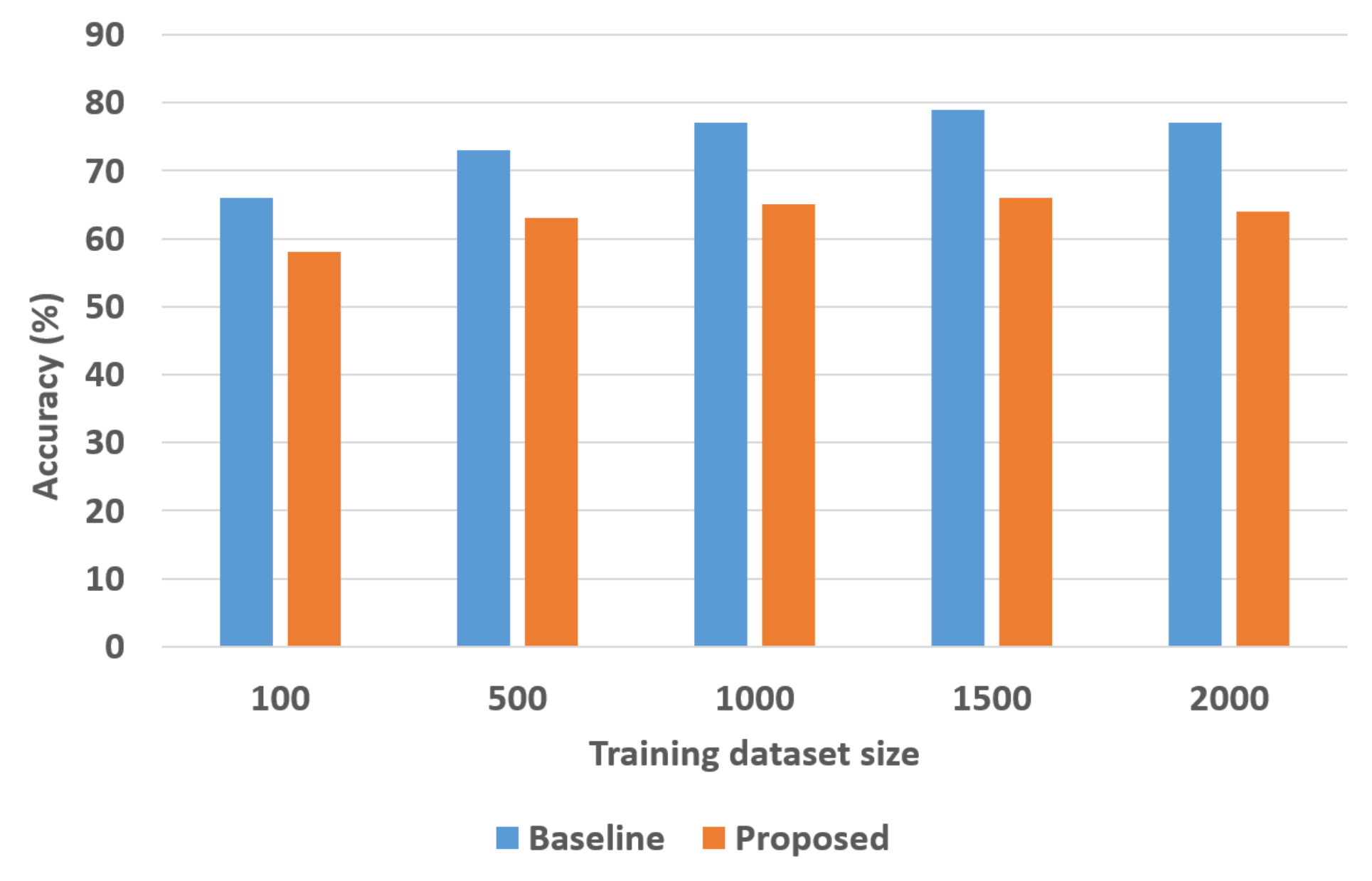}
  \vspace{-1em}
  \caption{Classification accuracy as a function of the number of
    training samples in the SS domain.}
  \label{fig:ss-classify}
\end{figure*}

\begin{figure*}[tbh]
  \centering
  \includegraphics[width=0.7\textwidth]{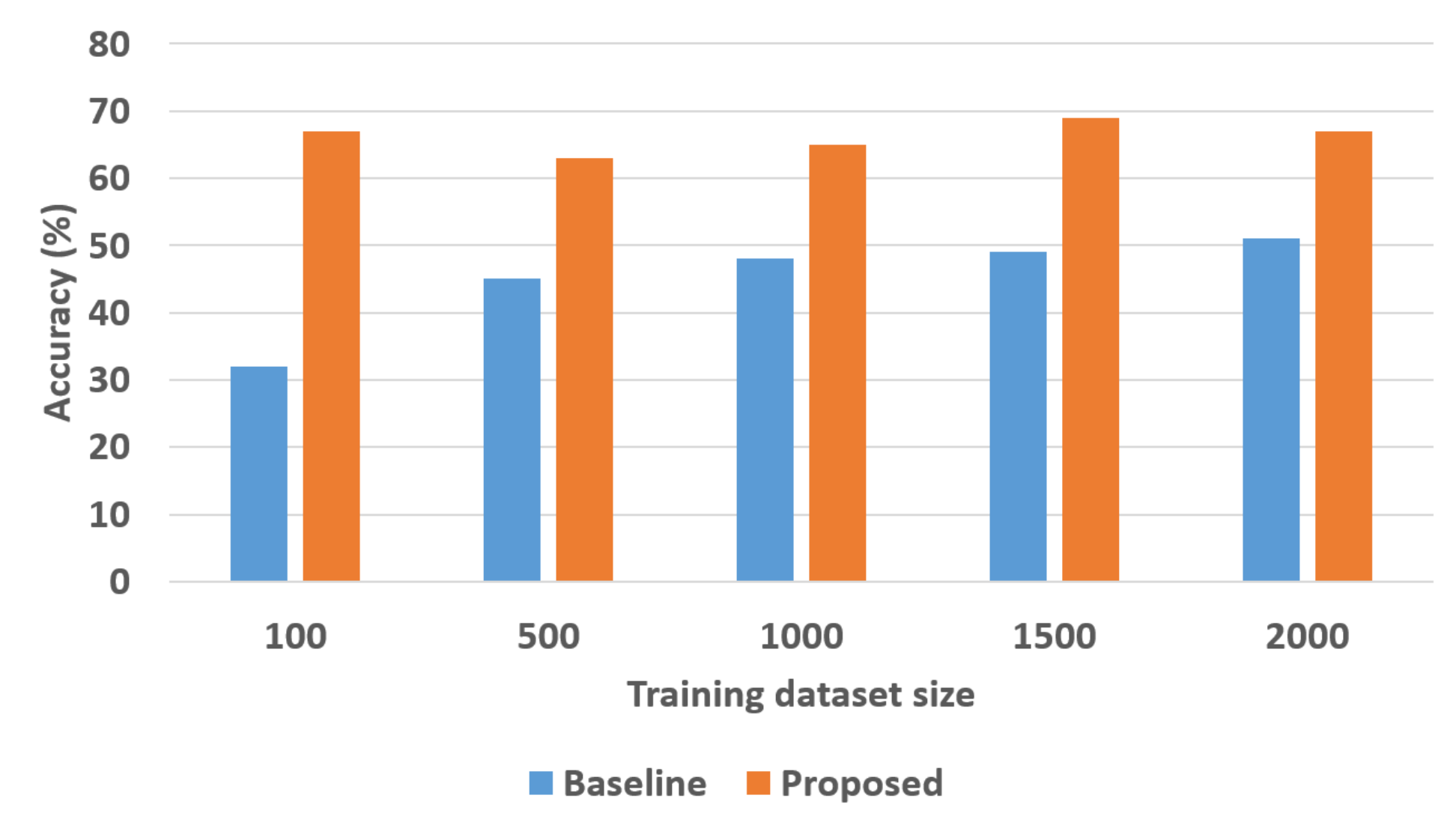}
  \vspace{-1em}
  \caption{VQA accuracy as a function of the number of training
    samples in the SS domain.}
  \label{fig:ss-vqa}
\end{figure*}

\begin{figure*}[tbh]
  \centering
  \includegraphics[width=0.7\textwidth]{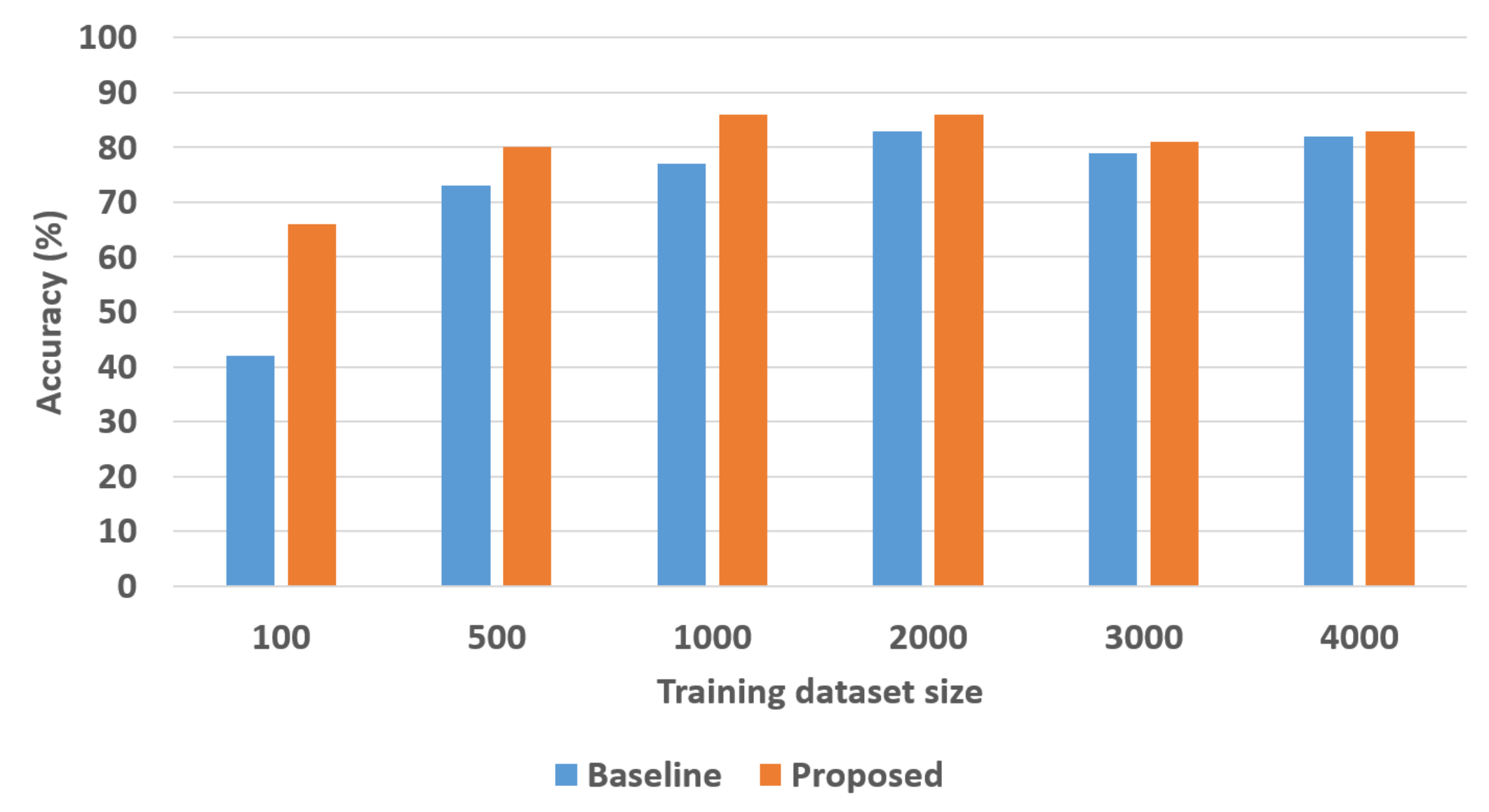}
  \vspace{-1em}
  \caption{Classification accuracy as a function of number of training
    samples in TS domain.}
  \label{fig:ts-classify}
\end{figure*}

\begin{figure*}[tbh]
  \centering
  \includegraphics[width=0.7\textwidth]{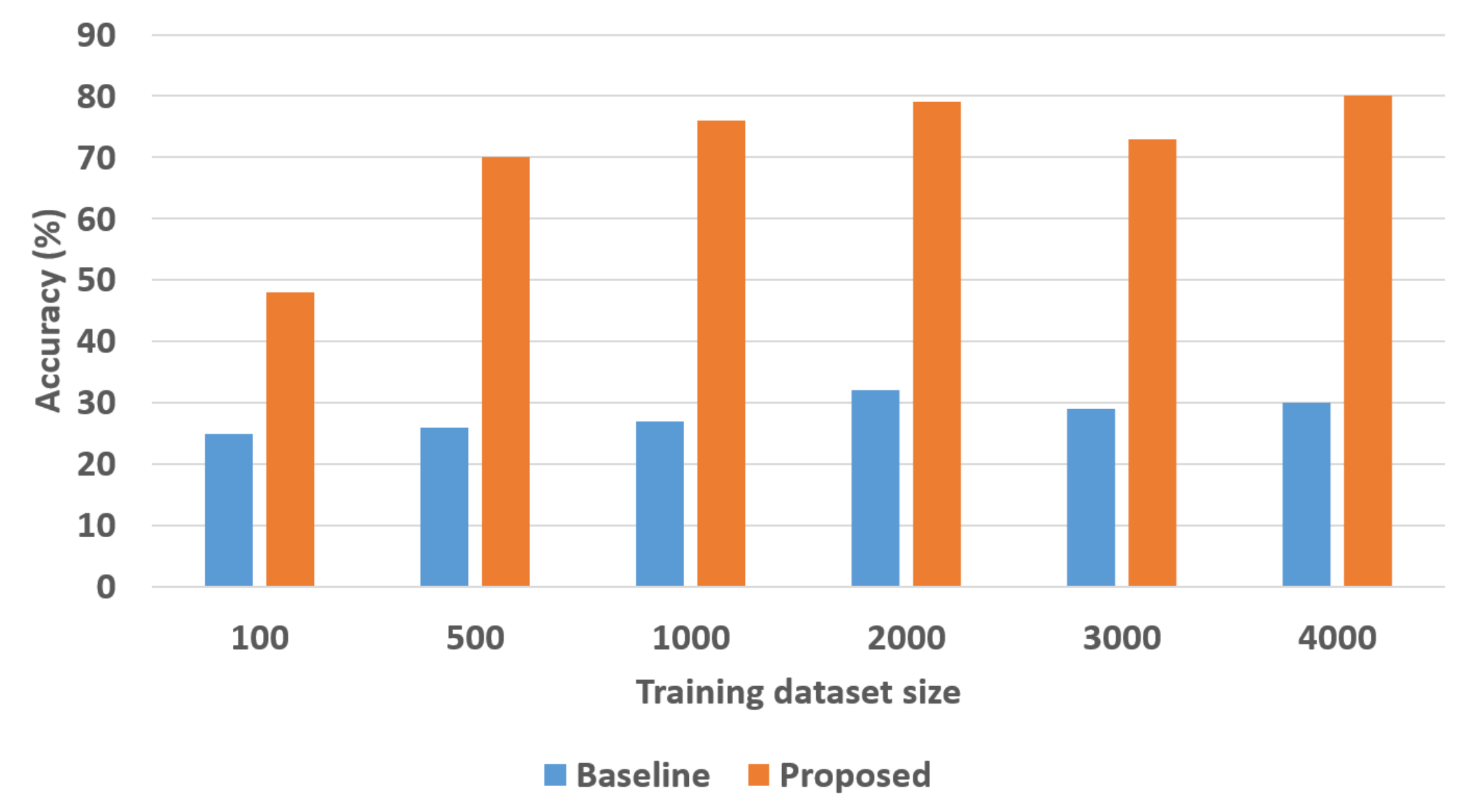}
  \vspace{-1em}
  \caption{VQA accuracy as a function of number of training samples in
    TS domain.}
  \label{fig:ts-vqa}
\end{figure*}

To quantitatively evaluate hypotheses $H1$ and $H2$, we ran
experimental trials in which we varied the size of the training
dataset. In these trials, the baseline performance was provided by a
CNN-RNN architecture, with the CNNs processing images to extract and
classify features, and the RNN providing answers to explanatory
questions. We repeated the trials $50$ times (choosing the training
set randomly each time) and the corresponding average results are
summarized in Figures~\ref{fig:ss-classify} and~\ref{fig:ss-vqa} for
the SS domain, and in Figures~\ref{fig:ts-classify}
and~\ref{fig:ts-vqa} for the TS domain.  We make some observations
based on these figures:
\begin{enumerate}
\item The classification performance of our architecture depends on
  the domain. In the relatively simpler SS domain, the baseline deep
  network architecture is at least as accurate as our architecture,
  even with a small training set---see Figure~\ref{fig:ss-classify}.
  This is because small differences in the position and arrangement of
  blocks (which could almost be considered as noise) influence the
  decision about stability. For instance, two arrangements of blocks
  that are almost identical end up receiving different ground truth
  labels for stability, and it is not possible to draft rules based on
  abstract image features to distinguish between these cases. The
  baseline deep network architecture, which generalizes from data, is
  observed to be more sensitive to these small changes than our
  architecture. Exploring the reason for this performance is an
  interesting direction for further research.

\item In the more complex TS domain, our architecture provides better
  classification accuracy than the baseline architecture based on just
  deep networks, especially when the size of the training set is
  small---see Figure~\ref{fig:ts-classify}. The classification
  accuracy increases with the size of the training set\footnote{We
    limit ourselves to training sets that are not too large in order
    to match the focus of our paper.}, but our architecture is always
  at least as accurate as the baseline architecture.

\item Our architecture is much more capable of answering explanatory
  questions about the classification decisions than the baseline
  architecture. When the answer provided by our architecture does not
  match the ground truth, we are able to examine why that decision was
  made. We were thus able to understand and explain the lower
  classification accuracy of our architecture in the SS domain. The
  baseline architecture does not provide this capability.

\item Unlike classification, the VQA performance of our architecture
  is much better than that of the baseline architecture in both
  domains. Also performance does not improve just by increasing the
  size of training set, even in simpler domains, e.g., see
  Figure~\ref{fig:ss-vqa}.  This is because VQA performance also
  depends on the complexity of the explanatory questions. For more
  complex domains, the improvement in VQA accuracy provided by our
  architecture is much more pronounced, e.g., see
  Figure~\ref{fig:ts-vqa}.
\end{enumerate}
We explored the statistical significance of the observed performance
by running paired t-tests. We observed that the VQA performance of the
proposed architecture was significantly better than that of the
baseline architecture; this is more pronounced in the TS domain that
is more complex than the SS domain. Also, although the baseline
architecture provides better classification performance in the SS
domain, the difference is not always statistically significant.

To further explore the observed results, we obtained a ``confidence
value'' from the logits layer of each CNN used to extract a feature
from the input image. For each CNN, the confidence value is the
largest probability assigned to any of the possible values of the
corresponding feature, i.e., it is the probability assigned to the
most likely value of the feature. These confidence values are
considered to be a measure of the network's confidence in the
corresponding features being a good representation of the image. We
trained a version of our architecture in which if the confidence value
for any feature was low, the image features were only used to revise
the decision tree (during training), or were processed using the
decision tree (during testing). In other words, features that do not
strongly capture the essence of the image are not used for
non-monotonic logical reasoning; the deep network architectures
provide much better generalization to noise. We hypothesized that this
approach would improve the accuracy of classification and question
answering, but it did not make any significant difference in our
experimental trials. We believe this is because the extracted features
were mostly good representations of the objects of interest in the
images. We thus did not use such networks (that compute the confidence
value) in any other experiments.

\subsection{\bf Experimental Results: Learn Axiom + VQA}
\label{sec:expresults-axiom-vqa}
\begin{figure*}[tbh]
  \centering
  \includegraphics[width=0.7\linewidth]{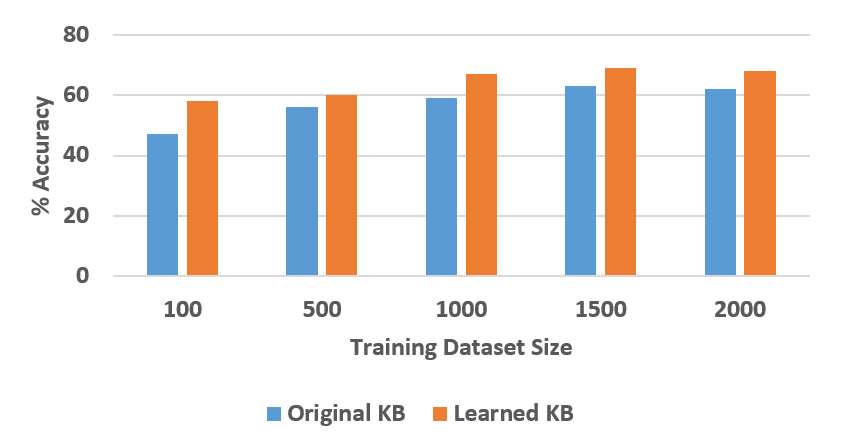}
  \vspace{-1em}
  \caption{Comparison of classification accuracy in the SS domain with
    and without axiom learning. In both cases, some axioms were missing
    from the knowledge base.}
  \label{fig:ss-axiom-classify}
\end{figure*}

\begin{figure*}[tbh]
  \centering
  \includegraphics[width=0.7\textwidth]{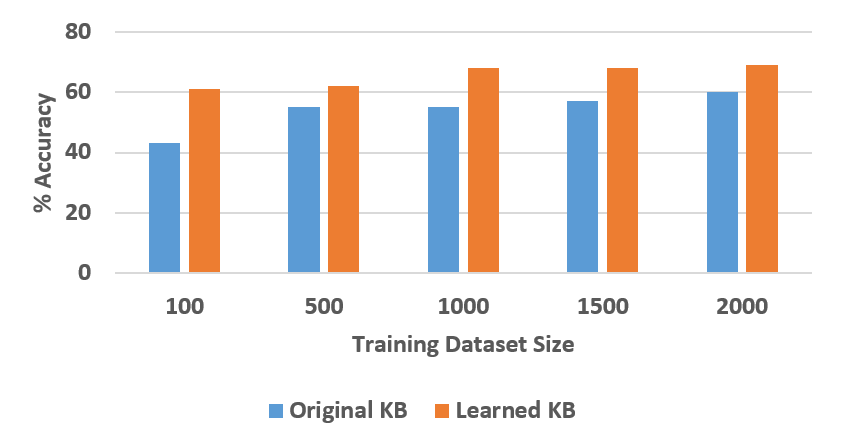}
  \vspace{-1em}
  \caption{Comparison of VQA accuracy in the SS domain with and
    without axiom learning. In both cases, some axioms were missing
    from the knowledge base.}
  \label{fig:ss-axiom-vqa}
\end{figure*}

\begin{figure*}[tbh]
  \centering
  \includegraphics[width=0.7\textwidth]{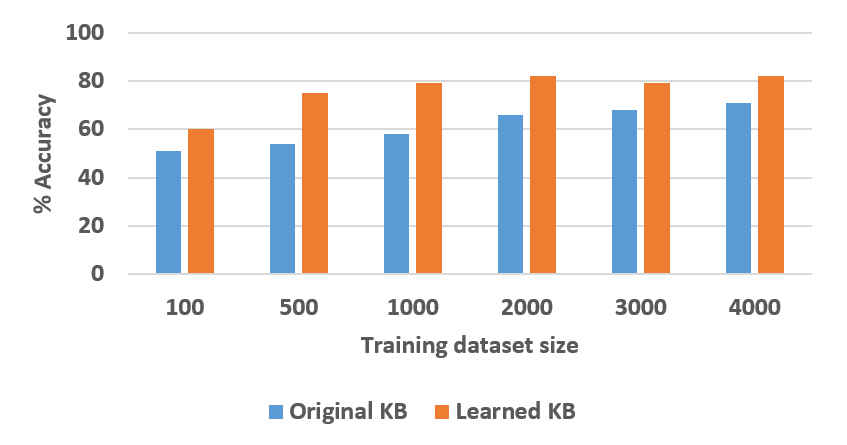}
  \vspace{-1em}
  \caption{Comparison of classification accuracy in the TS domain with
    and without axiom learning. In both cases, some axioms were missing
    from the knowledge base.}
  \label{fig:ts-axiom-classify}
\end{figure*}

\begin{figure*}[tbh]
  \centering
  \includegraphics[width=0.7\textwidth]{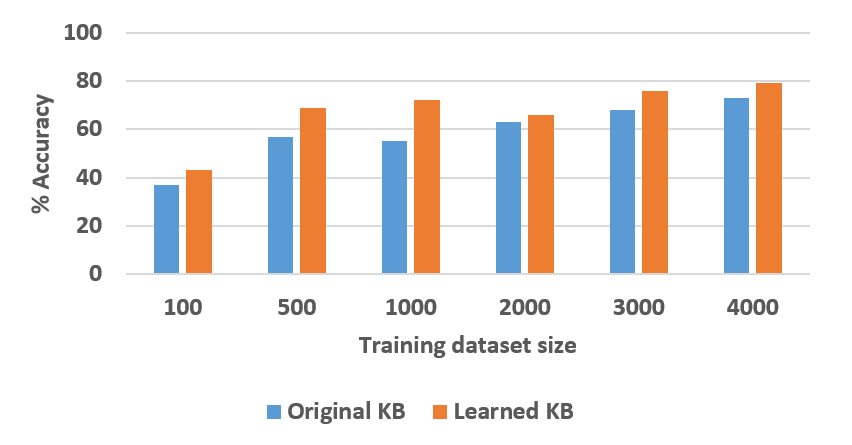}
  \vspace{-1em}
  \caption{Comparison of VQA accuracy in the TS domain with and
    without axiom learning. In both cases, some axioms were missing
    from the knowledge base.}
  \label{fig:ts-axiom-vqa}
\end{figure*}

Next, we experimentally evaluated the ability to learn axioms, and the
effect of such learning on the classification and VQA performance. For
the SS domain, we designed a version of the knowledge base with eight
axioms related to stability or instability of the structures. Out of
these, four were chosen (randomly) to be removed and we examined the
ability to learn these axioms, and the corresponding accuracy of
classification and VQA, as a function of the number of labeled
training examples (ranging from $100$ to $2000$). We repeated these
experiments $30$ times and the results (averaged over the $30$ trials)
are summarized in
Figures~\ref{fig:ss-axiom-classify}-\ref{fig:ss-axiom-vqa}. In the TS
domain, the methodology for experimental evaluation was the same.
However, since the domain was more complex, there were many more
axioms in the domain description (for classification and VQA); we also
had access to more labeled training examples. In each experimental
trial, a quarter of the available axioms were thus selected and
commented out, and the accuracy of classification and VQA were
evaluated with the number of labeled training examples varying from
$100$ to $4000$. The results averaged over $30$ such trials are
summarized in
Figures~\ref{fig:ts-axiom-classify}-\ref{fig:ts-axiom-vqa}.

In these figures, ``Original KB'' (depicted in blue) represents the
baseline with some axioms missing from the system description, e.g.,
four in the SS domain and one quarter of the axioms in the TS domain.
The results obtained by using the available labeled examples to learn
the axioms that are then used for classification and answering
explanatory questions about the scene, are shown as ``Learned KB'' in
orange. We observe that our approach supports incremental learning of
the domain axioms, and that using the learned axioms improves the
classification accuracy and the accuracy of answering explanatory
questions, in comparison with the baseline. This improvement was found
to be statistically significant using paired tests at $95\%$ level of
significance. These results support hypothesis $H3$.

\subsection{\bf Experimental Results: Learn Axiom + Plan}
\label{sec:expresults-axiom-plan}
Next, we experimentally evaluated the ability to learn axioms and the
effect of the learned axioms on planning, in the RA domain. The
simulated robot was equipped with domain knowledge for planning,
classification, and question answering. It uses this knowledge to
navigate through an office building, locate the intended recipient of
a message, deliver the message, detect and reason about objects in its
surroundings, and answer questions about the rooms it has visited.  As
stated in Section~\ref{sec:arch-plan}, we limit uncertainty in sensing
and actuation on robots to noise added in simulation. Average results
from $100$ trials indicates a VQA accuracy of $\approx 85\%$ after
training the architecture's components with just $500$ labeled images.
The domain knowledge includes learned axioms---the corresponding
experimental results and the planning performance are discussed later
in this section. We begin with an execution trace in this domain.

\noindent
\begin{execexample}\label{ex:ra-exec-example4}[Question Answering, RA
  Domain]
  {\rm Consider the scenario in the RA domain
    (Figure~\ref{fig:ra-domain}) in which the robot's goal was to
    deliver a message from John to Sally, and return to John to answer
    questions.
    \begin{itemize}
    \item The robot was initially in John's office. It computed a plan
      that comprises moving to Sally's office through the library and
      the kitchen, delivering the message to Sally, and returning to
      John's office through the same route to answer questions.
      
    \item During plan execution, the robot periodically takes images
      of the scenes in the domain,
      which are used for planning, classification and question
      answering.

    \item After returning to John's office, the robot and the human
      had an exchange about the plan constructed and executed, and the
      observations received. The exchange includes instances such as: \\
      \textbf{John's question:} ``is Sally's location cluttered?''\ \\
      \textbf{Robot's answer:} ``Yes''.\\
      When asked, the robot provides an \textbf{explanation} for this
      decision: ``Sally is in her office. Objects detected are Sally's
      chair, desk, and computer, and a cup, a large box, and a sofa.
      The room is cluttered because the cup, large box, and sofa are
      not usually in that room''.
    \end{itemize}
  }
\end{execexample}

The RA domain was also used to evaluate the effects of axiom learning.
There were four employees in offices in the simulated scenario, as
shown in Figure~\ref{fig:ra-domain}, and the robot was asked to find
particular individuals and deliver particular messages to them.
Employees are initially expected to be in their assigned workplace
(i.e., their office), and spend most of their time in these offices,
unless this default knowledge has been negated by other knowledge or
observations. This information is encoded as follows:
\begin{align}
  \nonumber
  &holds(loc(P, L), 0)~\leftarrow~ not~default\_negated(P,
  L), ~workplace(P, L)
\end{align}
where $workplace(P, L)$ specifies the default location of each person,
and $default\_negated(P, L)$ is used to encode that a particular
person may not be in their default location. These exceptions to the
defaults can be encoded as follows:
\begin{subequations}
  \label{eqn:ra-def-neg}
  \begin{align}
    &default\_negated(P, L)~\leftarrow~obs(loc(P, L1), true, I), ~L\neq L1 \\
    &default\_negated(P, L)~\leftarrow~obs(loc(P, L), false, I)
  \end{align}
\end{subequations}
Statement~\ref{eqn:ra-def-neg}(a) implies that the default assumption
should be ignored if the person in question is observed to be in a
location other than their workplace, and
Statement~\ref{eqn:ra-def-neg}(b) implies that a default assumption
should be ignored if the corresponding person is not observed in their
workplace. Including such default knowledge (and exceptions) in the
reasoning process allows the robot to compute better plans and execute
the plans more efficiently, e.g., when trying to deliver a message to
a particular person.  However, this knowledge may not be known in
advance, the existing knowledge may be inaccurate or change with time
(e.g., humans can move between the different places), or the
observations may be incorrect.  Our axiom learning approach was used
in this domain to acquire previously unknown information about the
default location of people and exceptions to these defaults. In all
the trials, the simulated robot was able to learn the previously
unknown axioms.


\begin{table*}[tbh]
  \centering
  \begin{tabular}{| c || c | c | c | c | c |} 
    \hline
    Axiom learning & Plans       & Actions     & Execution time & Planning time & Planning time\\ [0.5ex] 
                   & (per trial) & (per trial) & (per trial)    & (per trial)   & (per plan)\\ \hline \hline
    Before & 4 & 2.3 & 1.6 & 6.0 & 1.6 \\ \hline
    After  & 1 &  1  &  1  &  1  &  1 \\  [1ex] \hline
  \end{tabular}
  \caption{Planning performance in a scenario in the RA domain (see Figure~\ref{fig:ra-domain}) before and after axiom learning. Results averaged over $100$ paired trials indicate that reasoning with previously unknown axioms results in fewer plans with fewer actions in each trial, and significantly reduces the time taken to compute and execute the plans.}
  \label{tab:ra-axiom-learn}
\end{table*}

We then conducted $100$ paired trials to explore the effects of the
learned axioms on planning, with the corresponding results summarized
in Table~\ref{tab:ra-axiom-learn}. In each trial, we randomly chose a
particular goal and initial conditions, and measured planning
performance before and after the previously unknown axioms had been
learned and used for reasoning. Since the initial conditions are
chosen randomly, the object locations, the initial location of the
robot, and the goal, may vary significantly between trials. Under
these circumstances, it is not meaningful to average the results
obtained in the individual trials for performance measures such as
planning time and execution time. Instead, the results obtained
without including the learned axioms were computed as a ratio of the
results obtained after including the learned axioms; the numbers
reported in Table~\ref{tab:ra-axiom-learn} are the average of these
computed ratios. Before axiom learning, the robot often explored an
incorrect location (for a person) based on other considerations (e.g.,
distance to the room) and ended up having to replan. After the
previously unknown axioms were included in the reasoning process, the
robot went straight to the message recipient's most likely location,
which also happened to be the actual location of the recipient in many
trials. As a result, we observe a (statistically) significant
improvement in planning performance after the learned axioms are used
for reasoning. Note that in the absence of the learned axioms, the
robot computes four times as many plans taking six times as much time
in any given trial (on average) as when the learned axioms are
included in reasoning. Even the time taken to compute each plan (with
potentially multiple such plans computed in each trial) is
significantly higher in the absence of the learned axioms. This is
because the learned axioms enable the robot to eliminate irrelevant
paths in the transition diagram from further consideration. In a
similar manner, reasoning with intentional actions enables the robot
to significantly reduce the plan execution time by terminating or
revising existing plans when appropriate, especially in the context of
unexpected successes and failures. These results provide evidence in
support of hypothesis $H4$.


Finally, we conducted some initial proof of concept studies exploring
the use of our architecture on physical robots. We considered a robot
collaborating with a human to jointly describe structures of blocks on
a tabletop (similar to the SS domain described in this paper). We also
considered a mobile robot finding and moving objects to desired
locations in an indoor domain (similar to the RA domain). These
initial experiments provided some promising outcomes.  The robot was
able to provide answers to explanatory questions, compute and execute
plans to achieve goals, and learn previously unknown constraints. In
the future, we will conduct a detailed experimental analysis on robots
in different domains.

\section{Discussion and Future Work}
\label{sec:conclusions}
Visual question answering (VQA) combines challenges in computer
vision, natural language processing, and explainability in reasoning
and learning. Explanatory descriptions of decisions help identify
errors, and to design better algorithms and frameworks. In addition,
it helps improve trust in the use of reasoning and learning systems in
critical application domains.  State of the art algorithms for VQA are
based on deep networks and the corresponding learning algorithms.
Given their focus on generalizing across different domains, these
approaches are computationally expensive, require large training
datasets, and make it difficult to provide explanatory descriptions of
decisions. We instead focus on enabling reliable and efficient
operation in any given domain in which a large number of labeled
training examples may not be available. Inspired by research in
cognitive systems, our architecture tightly couples representation,
reasoning, and interactive learning, and exploits the complementary
strengths of deep learning, non-monotonic logical reasoning with
commonsense knowledge, and decision tree induction. Experimental
results on datasets of real world and simulated images indicate that
our architecture provides the following benefits in comparison with a
baseline architecture for VQA based on deep networks:
\begin{enumerate}
\item Better accuracy, sample efficiency, and time complexity on
  classification problems when the training dataset is small, and
  comparable accuracy with larger datasets while still using only a
  subset of these samples for training;

\item Ability to provide answers to explanatory questions about the
  scenes and the underlying decision making problems (e.g.,
  classification, planning);

\item Incremental learning of previously unknown domain constraints,
  whose use in reasoning improves the ability to answer explanatory
  questions; and

\item Ability to adapt the complementary strengths of non-monotonic
  logical reasoning with commonsense domain knowledge, inductive
  learning, and deep learning, to address decision-making (e.g.,
  planning) problems on a robot.
\end{enumerate}
Our architecture opens up multiple directions of future work, which
will address the limitations of existing work and significantly extend
the architecture's capabilities. We discuss some of these extensions
below:
\begin{enumerate}
\item The results reported in this paper are based on image datasets
  (simulated, real-world) chosen or constructed to mimic domains in
  which a large, labeled dataset is not readily available. One
  direction of future work is to explore the use of our architecture
  in other domains that provide datasets of increasing complexity,
  i.e., with a greater number of features and more complex explanatory
  questions. This exploration may require us to consider larger
  datasets, and to examine the trade-off between the size of the
  training dataset, the computational effort involved in processing
  such a dataset with many labeled examples, and the effort involved
  in encoding and reasoning with the relevant domain knowledge.
  
\item In our architecture, we have so far used variants of existing
  network structures as the deep network components (i.e., CNN, RNN).
  In the future, we will explore different deep network structures in
  our architecture, using the explanatory answers to further
  understand the internal representation of these network structures.
  Towards this objective, it would be particularly instructive to
  construct and explore deep networks and logic-based domain
  representations that provide similar behavior on a set of tasks, or
  provide different behavior when operating on the same dataset. As
  stated in the discussion in
  Section~\ref{sec:expresults-classify-vqa}, such an exploration may
  help us better understand (and improve) the design and use of deep
  network models for different applications.

\item This paper used VQA as a motivating problem to address key
  challenges in using deep networks in dynamic domains with limited
  labeled training examples. We also described the use of our
  architecture (with tightly-coupled reasoning and learning
  components) for planning on a simulated robot. In the future, we
  will combine this architecture with other architectures we have
  developed for knowledge representation, reasoning, and interactive
  learning in robotics~\cite{mohan:ACS18,mohan:JAIR19}. The long-term
  goal will be to support explainable reasoning and learning on a
  physical robot collaborating with humans in complex domains.
\end{enumerate}

\section*{Acknowledgments}
The authors thank Ales Leonardis for feedback on the architecture
described in this paper. 


\section*{Data Availability Statement}
The datasets generated or analyzed for this study, and the software
implementation of the architecture and algorithms, can be found in the
following online repository:
\url{https://github.com/hril230/masters_code}


\bibliographystyle{plain}
\bibliography{references}


\end{document}